\documentclass{fairmeta}

\usepackage[utf8]{inputenc}
\usepackage{amsmath,amssymb}
\usepackage{array}
\usepackage{colortbl}
\usepackage{adjustbox}
\usepackage{algorithm}
\usepackage{algorithmic}
\usepackage{listings}
\usepackage{xurl}
\usepackage{wrapfig}
\usepackage{needspace}
\usepackage{float}
\definecolor{OursRow}{RGB}{248,248,248}
\definecolor{EMGain}{RGB}{85,85,85}
\definecolor{EMPromptBody}{RGB}{248,248,248}
\newcommand{\bestgroup}[1]{\underline{#1}}
\newcommand{\bestoverall}[1]{\textbf{#1}}
\newcommand{\gaincell}[2]{\shortstack{#1\\{\scriptsize\textcolor{EMGain}{(#2)}}}}
\newcommand{\mypar}[1]{\par\vspace{1mm}\textbf{#1}}
\renewcommand{\paragraph}[1]{\par\vspace{1.25mm}\noindent\textbf{#1}\ }

\titlespacing*{\section}{0pt}{2.5ex plus 0.5ex minus 0.2ex}{1.2ex plus 0.2ex}
\titlespacing*{\subsection}{0pt}{2ex plus 0.5ex minus 0.2ex}{0.8ex plus 0.2ex}
\setlabdisplayname{OmniAI Group of ZJU ACES Lab}
\setuniversityname{Zhejiang University}

\title{EmbodiedMemory-Bench: Benchmarking Embodied Memory for Long-Horizon Embodied Tasks}
\author[1]{Lizhou Liang}
\author[2]{Xinyu Zhong}
\author[1]{Miao Pan}
\author[1]{Xiaohe Zhou}
\author[1]{Xuanyu Liu}
\author[1]{Qinfeng Li}
\author[3]{Peng Li}
\author[1]{Jintao Chen}
\author[1]{Xuhong Zhang}
\author[1]{Wenqi Zhang}
\affiliation[1]{Zhejiang University}
\affiliation[2]{Central South University}
\affiliation[3]{Institute of Software, Chinese Academy of Sciences}

\abstract{
Long-horizon embodied interaction requires agents to retain and continually update information about the environment as they observe, act, and encounter change. Yet current agents struggle to maintain such memory reliably. Our analysis traces this limitation to four key deficiencies: weak fine-grained visual memory, unreliable dynamic world-state tracking, failing to record world state revealed by interaction outcomes, and limited generalization from prior experience. However, existing benchmarks do not directly assess these memory capabilities during long-horizon embodied interaction. To address this gap, we introduce \textbf{EmbodiedMemory-Bench (EMem-Bench)}, comprising 2,554 interactive episodes across four task families. EMem-Bench requires agents to build and update memory from interaction history, then use it to complete a later task by acting in the environment. We further present \textbf{Embodied-Memorizer (EMem)}, an external memory system that organizes embodied experience into spatial, event, and scene memories. We also train EMem-8B, an 8B policy that manages and uses these memories. We evaluate a diverse range of open-source and proprietary MLLMs and representative multimodal memory systems. Results show that current models remain weak and uneven across the four challenges. Under matched backbones, EMem achieves the best overall performance among the evaluated memory systems and improves both open-source and proprietary models, while EMem-8B further improves over its backbone.
Project~page: \url{https://zju-omniai.github.io/EmbodiedMemoryBench/}.
}

\hypersetup{
  bookmarksdepth=2,
  pdftitle={EmbodiedMemory-Bench: Benchmarking Embodied Memory for Long-Horizon Embodied Tasks},
  pdfsubject={Embodied memory benchmark and Embodied-Memorizer}
}

\begin{document}
\thispagestyle{firstheader}
\maketitle
\pagestyle{empty}

\section{Introduction}

Recent rapid advances in language models have empowered LLM agents to autonomously complete complex, long-horizon tasks in digital environments, often spanning hours or even days, such as developing large-scale software from scratch or conducting online research and writing reports. A key capability underlying this progress is long-term memory, which involves retaining interaction histories, retrieving past information, and reusing experience across tasks.

\begin{figure}[!htbp]
\centering
\includegraphics[width=0.95\linewidth]{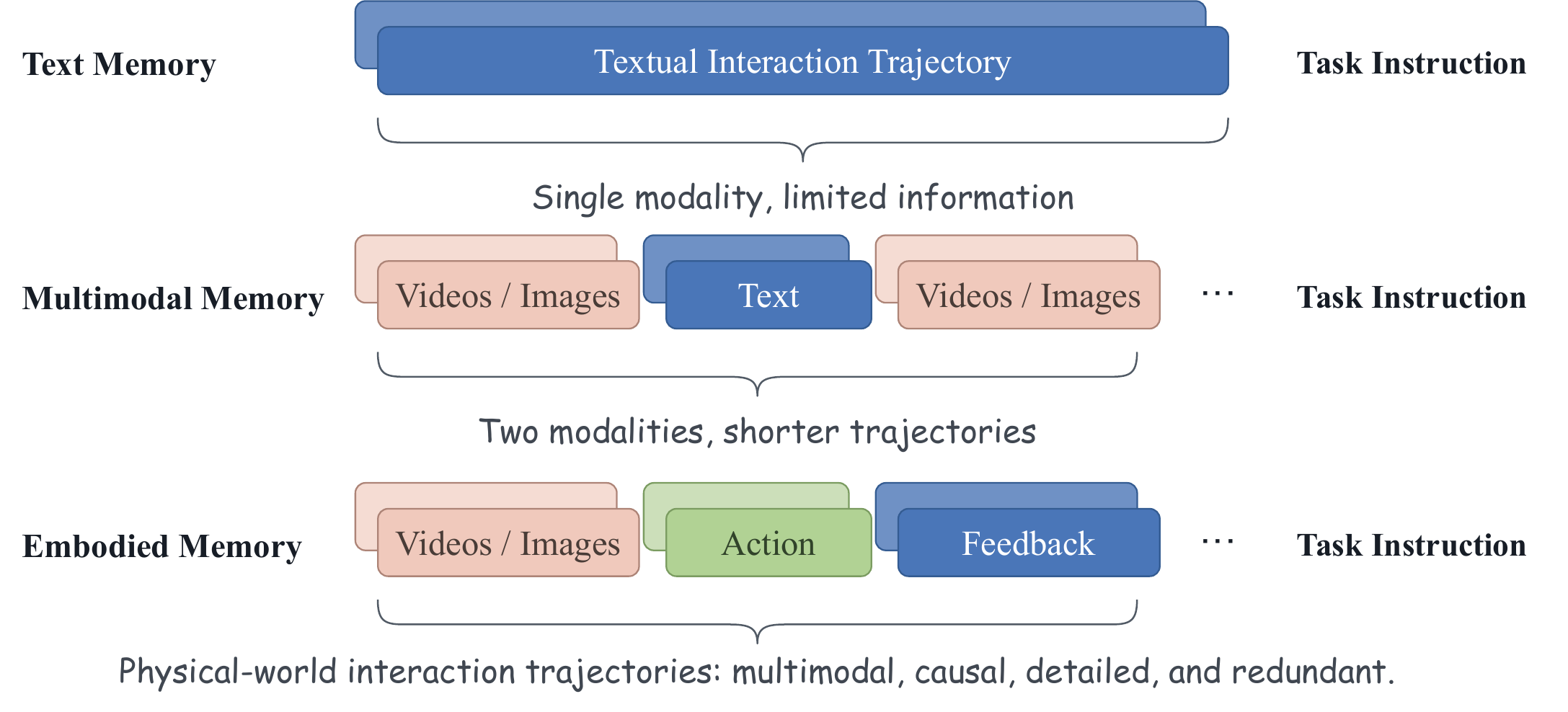}
\caption{Comparison of text-centric, multimodal, and embodied memory. Embodied memory connects visual observations, actions, and feedback into causal interaction histories that support subsequent tasks in a changing physical world.}
\label{fig:memory-comparison}
\end{figure}

When agents step out of \textbf{text-centric digital environments} into the \textbf{real 3D physical world}, however, the demands on memory become substantially more complex. Text-centric memory primarily retains linguistic information, while multimodal memory extends this to images and videos. As illustrated in Figure~\ref{fig:memory-comparison}, embodied memory must further preserve the \textbf{causal connections between observations, actions, and feedback}: actions change the environment, and their outcomes reveal world states and constraints that may not be directly visible. These interactions unfold over long horizons in environments subject to disturbance and change~\citep{zhang2025embodiedreasonersynergizingvisualsearch}. Embodied agents must therefore process \textbf{cross-modal}, \textbf{lengthy}, and \textbf{noisy} interaction histories, extracting useful context and organizing it into \textbf{concise memory} that supports reasoning and subsequent physical interaction.

Likewise, embodied memory is an indispensable capability for long-horizon interaction in the physical world: an embodied agent that cannot remember where objects were observed, how the environment has changed, or what actions it has previously taken is inevitably reduced to short-sighted behavior driven by immediate perception. 

\begin{wrapfigure}{r}{0.45\textwidth}
\vspace{0pt}
\centering
\includegraphics[width=\linewidth]{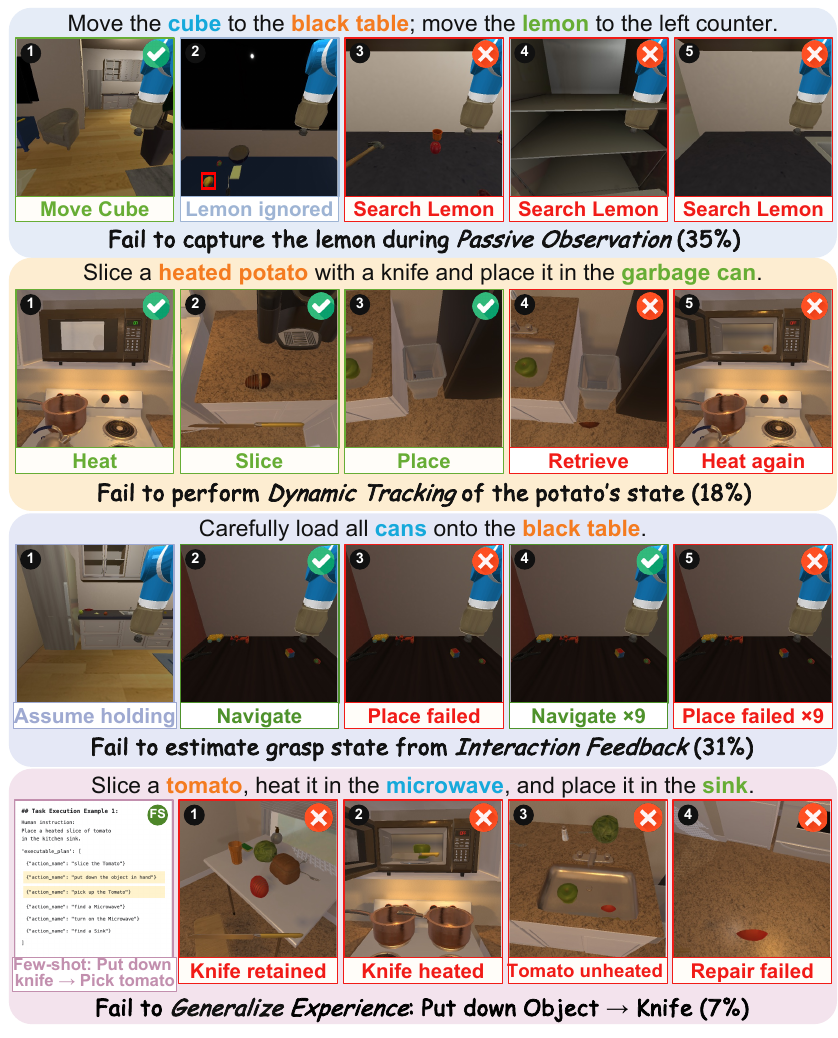}
\caption{Primary errors in 100 failed EmbodiedBench trajectories from Gemini-3-Flash and Qwen3-VL-32B on EB-ALF and EB-Hab. The four embodied-memory categories account for 91\% of failures. Percentages in parentheses indicate the proportion of failures attributed to each error category.}
\label{fig:embodiedbench-failure}
\end{wrapfigure}

Recent embodied benchmarks~\citep{yang2025embodiedbench} show that even the most advanced MLLMs struggle with long-horizon embodied tasks. To investigate this limitation in depth, we manually inspect 100 failed trajectories of Gemini-3-Flash and Qwen3-VL-32B on EmbodiedBench's EB-ALF and EB-Hab. As shown in Figure~\ref{fig:embodiedbench-failure}, most failures fall into four categories: \textbf{forgetting fine-grained visual cues} (35\%), \textbf{overlooking changes in the environment} (18\%), \textbf{failing to record world state revealed by interaction outcomes} (31\%), and \textbf{failing to generalize from prior experience} (7\%). For long-horizon tasks, these four types of failures are so frequent and pervasive (about 91\% in total) that even advanced MLLMs and embodied models are not immune to them.
This four typical failures correspond to four essential embodied-memory capabilities in long-horizon interaction: \textbf{(1) Fine-grained visual memory.} Although future tasks are not known in advance, an embodied agent must remember the appearances and locations of the objects it has seen, as any of them may be needed later. \textbf{(2) Dynamic world-state tracking.} In real-world environments, the states of objects may be altered by humans or other robots, so the agent must continually record the latest state of each object. \textbf{(3) Recording world state revealed by interaction outcomes.} Failed grasps, locked cabinets, and unusable receptacles reveal state unavailable to vision; this state must likewise be remembered and constrain later actions. \textbf{(4) Experience generalization memory.} Successes and failures from earlier tasks, together with physical regularities observed along the way, must also be retained and transferred to new tasks.

In this paper, we introduce \textbf{EmbodiedMemory-Bench (EMem-Bench)}, which contains 2,554 carefully designed episodes. Each episode consists of a temporally ordered history of multimodal interactions with the environment $\mathcal{H}=\{o^{\mathrm{h}}_1,\dots,o^{\mathrm{h}}_N\}$, a target task $\mathcal{T}$, and a feasible action space $\mathcal{A}$. The embodied agent must comprehend the given history $\mathcal{H}$, implicitly or explicitly extract the information required by $\mathcal{T}$, and then interact with the environment step by step, generating its own interaction trajectory ($o^{\mathrm{e}}_{1:t}$) until the task is completed: $a_t \sim \pi\!\left(\cdot \mid \mathcal{H}, \mathcal{T}, o^{\mathrm{e}}_{1:t}, a_{1:t-1}\right)$.

\begin{figure}[!htbp]
\centering
\includegraphics[width=0.9\textwidth]{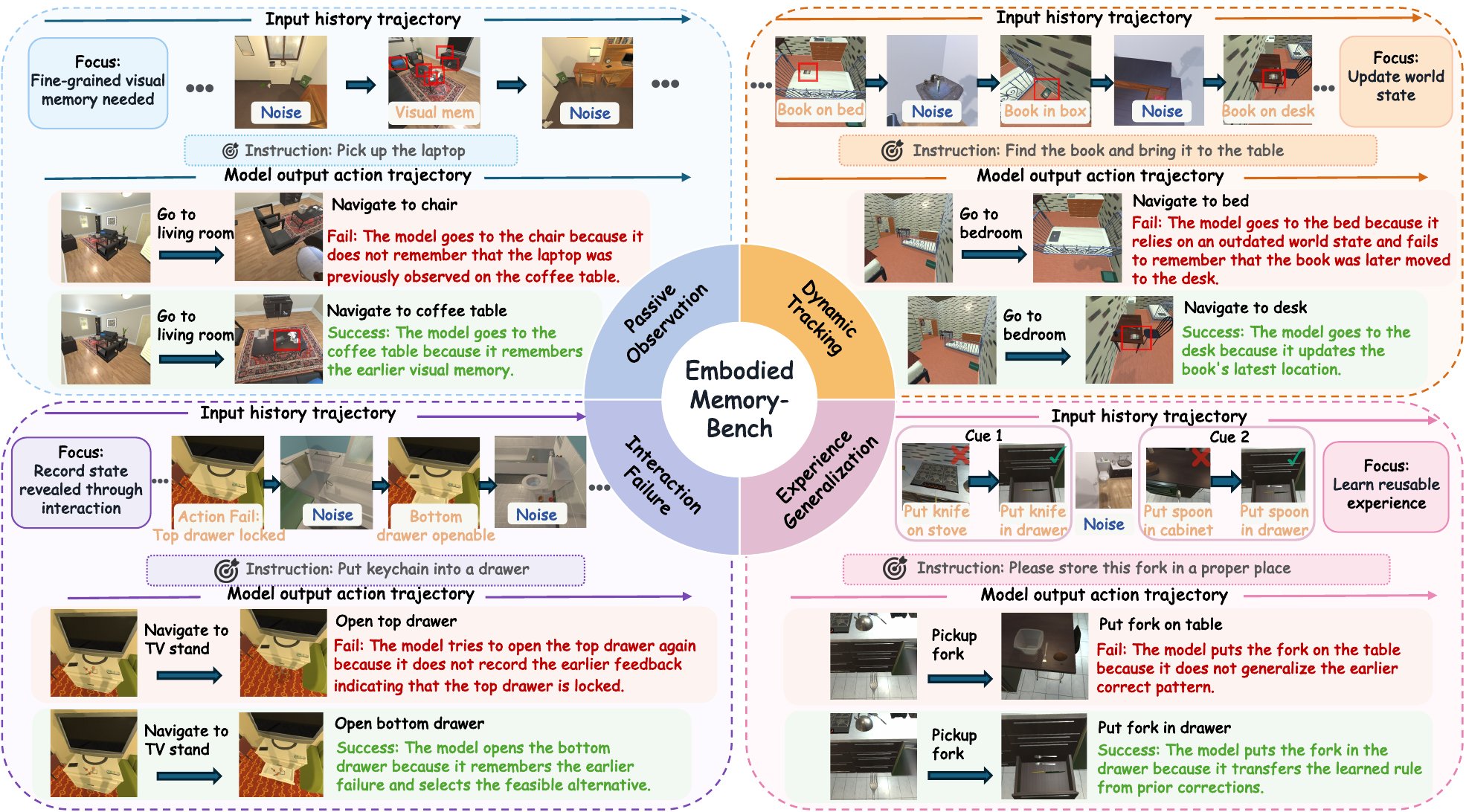}
\caption{Overview of EMem-Bench. Representative examples cover fine-grained visual memory, dynamic world-state tracking, recording world state through interaction, and experience generalization.}
\label{fig:benchmark-overview}
\end{figure}

We design each episode by varying the interaction history ($\mathcal{H}$) and the target task ($\mathcal{T}$). All tasks are organized into four sub-task families: \textbf{Passive Observation} tests fine-grained visual memory in object-dense views; \textbf{Dynamic Tracking} tests whether the agent remembers changes in object states within the environment; \textbf{Interaction Failure} tests whether world state revealed by interaction outcomes is recorded; and \textbf{Experience Generalization} tests whether regularities learned from recurring experiences transfer to new task scenarios. Figure~\ref{fig:benchmark-overview} illustrates the four task families.

EMem-Bench differs from current general embodied benchmarks, which either adopt single-turn question-answering task or instruct the agent to explore the environment and execute a given task without any provided context. In contrast, our memory-centric benchmark focuses on an embodied agent's ability to exploit memory: given offline interaction trajectories, the agent must consolidate them into memory, reason over it, and translate it into effective actions in the environment. Table~\ref{tab:positioning} summarizes this comparison.

Using EMem-Bench, we evaluate 16 leading open-source and proprietary MLLMs together with representative multimodal memory systems. The strongest proprietary model, Gemini-3-Flash, reaches only 64.2\% average SR, while all but one of the evaluated open-source models score below 45\%. Among 400 sampled failures, 61.3\% begin when the model acts on a memory contradicted by its interaction history. These results show that even strong models cannot consistently maintain an accurate memory of the environment they interact with, and that open-source models exhibit large differences across the four task families.   

In addition, we introduce a simple yet effective embodied memory system, \textbf{Embodied-Memorizer (EMem)}, as a baseline, which organizes multimodal experience through \textbf{scene}, \textbf{spatial}, and \textbf{event memories}, together with an 8B policy trained to write, retrieve, and use these memories. Without task-specific training, EMem improves average SR by 16.4 points on the open-source Mistral-Small-3.1-24B and by 23.6 points on the proprietary GPT-5.4-mini; the trained EMem-8B policy improves over Qwen3-VL-8B by 20.6 points.
Our contributions are as follows:
\begin{itemize}
    \item We identify four typical errors in embodied interaction: forgetting fine-grained visual cues, overlooking environmental changes, failing to record world state revealed by interaction outcomes, and failing to generalize from prior experience---revealing that current MLLMs lack embodied memory required for long-horizon tasks.
    \item We introduce EMem-Bench, a memory-centric benchmark with 2,554 episodes across four task families. Agents must extract information from a given multimodal interaction history, construct the relevant memory, and interact with the environment for task completion.
    \item We systematically evaluate 16 open-source and proprietary MLLMs together with representative multimodal memory systems, revealing persistent limitations across the four memory types. We further present an embodied memory system that organizes interaction trajectories into scene, spatial, and event memories and improves both open-source and proprietary models.
\end{itemize}

\section{Related Work}

\begin{table}[tbp]
\setlength{\tabcolsep}{1.5pt}
\renewcommand{\arraystretch}{1.1}
\centering
{\small
\begin{tabular*}{\linewidth}{@{\extracolsep{\fill}}lccccc@{}}
\toprule
\textbf{Benchmark} & \textbf{Scenario} & \textbf{\mbox{Vis.}} & \textbf{\mbox{Dyn.}} & \textbf{\mbox{Int.}} & \textbf{\mbox{Exp.}} \\
\midrule
\mbox{MemoryAgentBench} & LLM Agent & -- & -- & -- & Y \\
Evo-Memory & LLM Agent & -- & -- & -- & Y \\
WorldMemArena & Multimodal & Y & Y & -- & Y \\
EmbodiedBench & Embodied & -- & -- & -- & -- \\
FindingDory & Embodied & Y & -- & -- & -- \\
LMEE-Bench & Embodied & Y & -- & -- & -- \\
SpaMEM & Embodied & Y & Y & -- & -- \\
WorldLines & Embodied & -- & Y & -- & -- \\
\midrule
\textbf{EMem-Bench} & Embodied & Y & Y & Y & Y \\
\bottomrule
\end{tabular*}}
\caption{Comparison with existing benchmarks. Vis., Dyn., Int., and Exp. denote visual memory, dynamic tracking, interaction-derived state, and experience generalization, respectively. Y marks explicit evaluation.}
\label{tab:positioning}
\end{table}

\mypar{Embodied-agent benchmarks.}
EmbodiedBench evaluates vision-driven agents across action levels. FindingDory and LMEE-Bench use visual history for navigation \citep{yadav2025findingdory,wang2026lmee}; SpaMEM tests spatial-state revision \citep{liao2026spamem}; and WorldLines evaluates state question answering and planning over multi-day traces \citep{zhang2026worldlines}. These benchmarks introduce memory within individual tasks, whereas EMem-Bench evaluates embodied memory jointly across interactive episodes.

\mypar{Agent memory.}
LoCoMo and LongMemEval evaluate memory over multi-session histories \citep{maharana2024locomo,wu2025longmemeval}. MemoryAgentBench studies retrieval and test-time learning \citep{hu2025memoryagentbench}; Evo-Memory studies experience reuse \citep{wei2025evomemory}; and WorldMemArena evaluates memory writing, maintenance, retrieval, and use \citep{liu2026worldmemarena}. Unlike EMem-Bench, these benchmarks do not require remembered information to guide subsequent embodied interaction. Additional related work is discussed in Appendix~\ref{app:extended-related-work}.

\section{EMem-Bench}
\label{sec:benchmark}

\subsection{Task Definition}
EMem-Bench evaluates whether embodied agents can maintain a world state over long-term interaction. Each episode is defined by a temporally ordered history of multimodal interactions $\mathcal{H}=\{o^{\mathrm{h}}_1,\ldots,o^{\mathrm{h}}_N\}$, a target task $\mathcal{T}$, and a feasible action space $\mathcal{A}$. The history contains memory evidence needed for the target task. After receiving $\mathcal{T}$, the model must retrieve the relevant evidence from $\mathcal{H}$ and use it to complete the task through continued interaction with the environment. At execution step $t$, the model selects an action according to
\begin{equation}
a_t \sim \pi\!\left(\cdot \mid \mathcal{H},\mathcal{T},o^{\mathrm{e}}_{1:t},a_{1:t-1}\right),
\qquad a_t \in \mathcal{A},
\end{equation}
where $o^{\mathrm{e}}_{1:t}$ denotes the observations received during task execution. An episode is successful when the resulting interaction trajectory reaches an environment state that satisfies $\mathcal{T}$. The benchmark is organized into four task families.

\IfFileExists{plots/fig_2.pdf}{%
\begin{figure}[!htbp]
\centering
\includegraphics[width=0.9\textwidth]{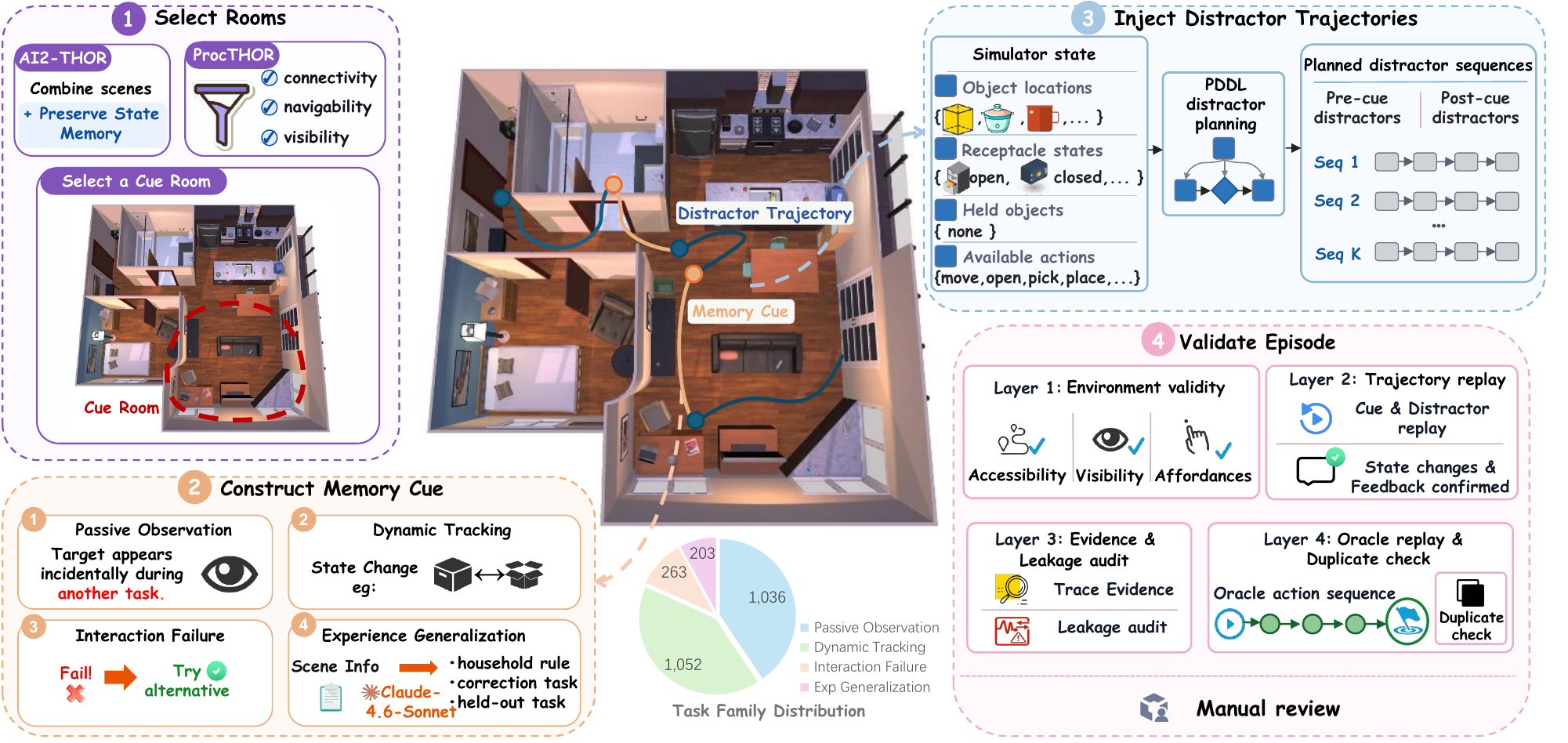}
\caption{Construction pipeline of EMem-Bench. Each episode combines a grounded scene, a task-specific memory cue, and distractor trajectories, followed by automated validation and manual review.}
\label{fig:construction-pipeline}
\end{figure}
}{}

\subsection{Task Families}

The four task families primarily differ in how $\mathcal{H}$ and $\mathcal{T}$ are constructed: $\mathcal{H}$ determines what type of information must be remembered, while $\mathcal{T}$ determines how that information must be used during later interaction.

\mypar{Passive Observation.}
This family evaluates whether an agent can build fine-grained visual memory in an object-dense scene. The interaction history contains a task trajectory in an object-dense scene, during which the later target object appears. The subsequent task asks the agent to find that object. In Figure~\ref{fig:benchmark-overview}, the laptop appears among several objects on the coffee table; navigating to the chair instead shows that the agent failed to retain its location.

\mypar{Dynamic Tracking.}
This family evaluates whether an agent can keep its world state current as object locations and states change over time. The interaction history records an object before and after its location or state changes. The subsequent task requires the agent to act on the latest state rather than an outdated observation. In Figure~\ref{fig:benchmark-overview}, the book moves from the bed through the box to the desk amid distractor observations; returning to the bed shows that the agent failed to update the world state stored in memory.

\mypar{Interaction Failure.}
This family evaluates whether an agent can retain world state revealed by interaction outcomes. The interaction history contains an attempted physical action and its failure feedback, which reveals a visually inaccessible state such as a locked drawer or blocked receptacle. The subsequent task presents several possible interaction targets and requires the agent to avoid the one shown to be unusable and choose a viable alternative. In Figure~\ref{fig:benchmark-overview}, a failed attempt reveals that the top drawer is locked; trying it again instead of the usable bottom drawer shows that the agent failed to record the locked state revealed by the interaction.

\mypar{Experience Generalization.}
This family evaluates whether an agent can transfer a regularity learned from past events to a new object. The interaction history contains several cases involving different objects, each showing an initially incorrect action and a correction. The subsequent task introduces a new object from the same category that did not appear in the histories and requires the agent to apply their shared regularity. In Figure~\ref{fig:benchmark-overview}, corrections place a knife and a spoon in a drawer before the agent encounters a fork; placing the fork on the table shows that the shared placement regularity did not transfer to the new object.

\subsection{Interaction-History Construction}

We construct the interaction history provided to the model in four stages: selecting a multi-room environment, creating a task-relevant memory cue, inserting distractor trajectories, and validating the resulting trajectory. Figure~\ref{fig:construction-pipeline} summarizes how this model input is constructed.

\mypar{Selecting rooms.}
AI2-THOR~\citep{kolve2017ai2thor} scenes contain only a single room, so we compose scenes of different room types into a virtual home and extend the action space with \texttt{LeaveRoom}, \texttt{EnterRoom}, and \texttt{MoveToRoom}. The state of room is restored when the agent re-enters it. ProcTHOR~\citep{deitke2022procthor} already provides multi-room homes; we filter them for room connectivity, navigable space, object visibility, and executable interactions. From each retained environment, we designate a cue room that supports the required memory relation and use the remaining connected rooms for unrelated activity and the later target task.

\mypar{Constructing the memory cue.}
We define a memory cue as an interaction trajectory containing the information that the agent must remember to complete the target task. We construct a different memory cue for each task family. For Passive Observation, we select an object that is visible in the agent's current view, then use it as the memory cue for a later retrieval task. For Dynamic Tracking, we select an object, change its location or state in the simulator, and make the later task depend on the updated state. For Interaction Failure, we execute a failed action whose feedback reveals an object state, and make the later task depend on that state. For Experience Generalization, Claude-4.6-Sonnet~\citep{anthropic2026sonnet46} receives structured scene information and outputs a structured experience containing an underlying regularity, several correction cases, and a target task. We then ground the generated experience in concrete scene objects and execute the resulting interaction trajectory in the simulator.

\mypar{Injecting distractor trajectories.}
A PDDL planner~\citep{aeronautiques1998pddl} generates unrelated trajectories before and after the memory cue based on the current scene and agent state. These trajectories vary where the cue appears in the episode and place other rooms and tasks between the cue and the target task. We discard any trajectory that touches critical objects, reveals the answer, changes the state required by the target task, or cannot be executed.

\mypar{Validating and reviewing each episode.}
Automated checks ensure that the scene is navigable, the required interactions can be executed, and the target task can be completed. They also verify that the decisive evidence appears in the earlier experience without being revealed in the distractor trajectories. We then manually inspect every candidate that passes these checks and remove episodes with ambiguous instructions, implausible trajectories, or cues that do not uniquely support the intended action. This process rejected 143 episodes; all 2,554 retained episodes pass both automated execution checks and manual screening.

\Needspace{4\baselineskip}
\subsection{Benchmark Statistics and Evaluation}

\begin{wrapfigure}{R}{0.45\textwidth}
\vspace{0pt}
\centering
\includegraphics[width=\linewidth]{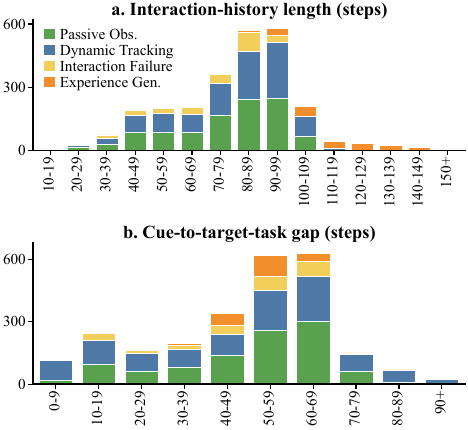}
\caption{Distributions of interaction-history length and cue-to-target-task distance.}
\label{fig:benchmark-difficulty}
\end{wrapfigure}

EMem-Bench contains 2,554 episodes: 1,036 Passive Observation, 1,052 Dynamic Tracking, 263 Interaction Failure, and 203 Experience Generalization. Across all four families, the benchmark spans 1,118 scenes, 125 visible object types, 83 target types, and 33 receptacle types. Figure~\ref{fig:benchmark-difficulty} shows the interaction-history length and task composition together with cue-to-target-task distance. Visible-object density at the cue and room/scene transitions are reported in the appendix.

\mypar{Evaluation metrics.}
Predicted action sequences are executed in the simulator, and an episode is successful only when the resulting environment reaches the target terminal state. We report success rate (SR) for each task family and use their equally weighted mean as the overall score, preventing the two larger families from dominating the evaluation. Full benchmark statistics are provided in the appendix.

We additionally report Error Recurrence Rate (ERR). For task family $f$ with $N_f$ episodes,
\begin{equation}
\mathrm{ERR}_f = \frac{1}{N_f}\sum_{i=1}^{N_f}
\min\!\left(1,\frac{E_i}{\max(1,K_i)}\right).
\end{equation}
where $E_i$ counts distinct error recurrences in episode $i$ and $K_i$ denotes its memory-dependent decision points. ERR is triggered by actions that use stale locations, repeat known failed interactions, or violate learned regularities.

\section{Our Method: Embodied-Memorizer}
\label{sec:method}

\begin{wrapfigure}{r}{0.45\textwidth}
\vspace{0pt}
\centering
\includegraphics[width=\linewidth]{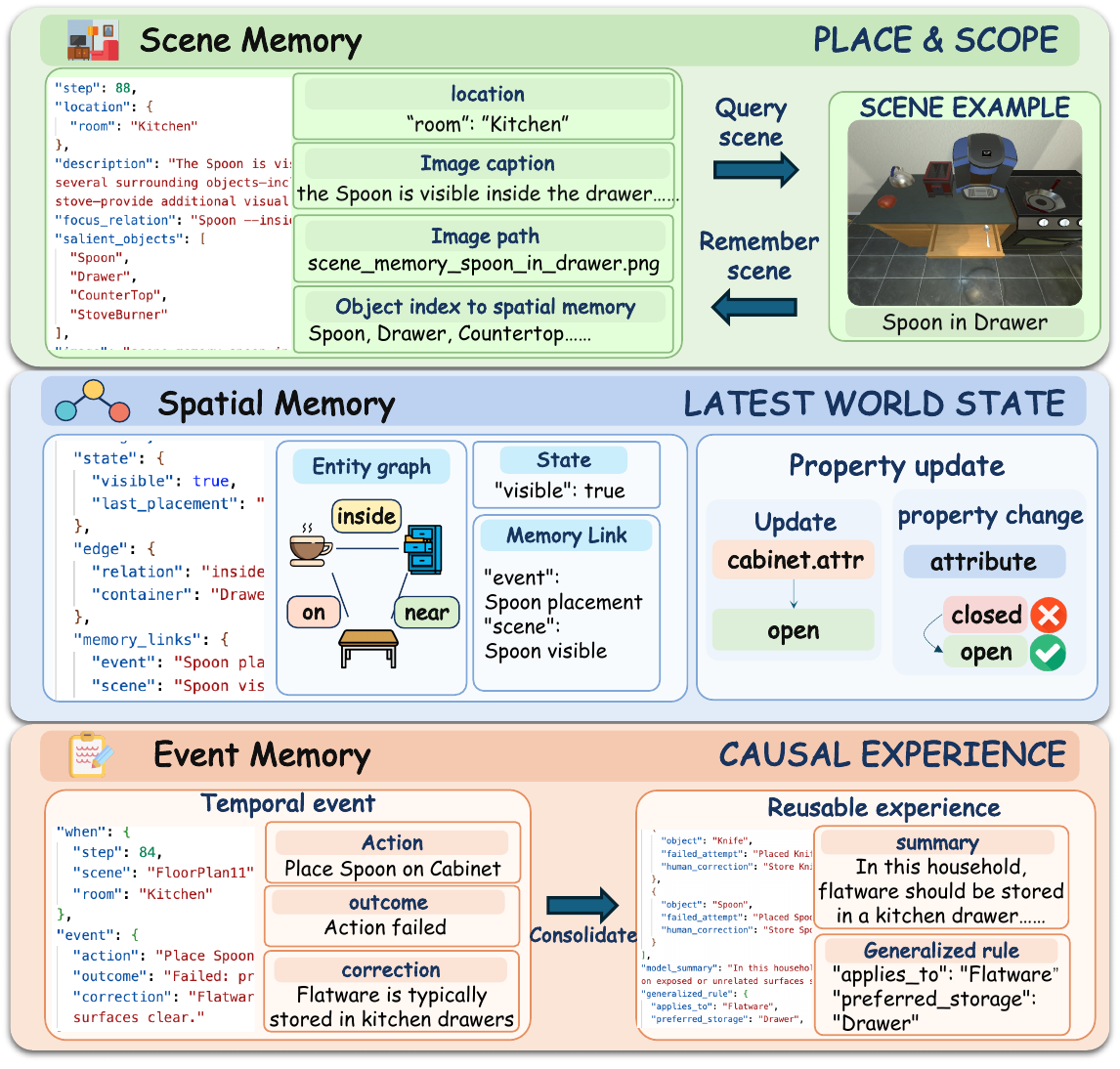}
\caption{EMem architecture. Scene memory stores visual context, spatial memory maintains current object states, and event memory records interactions and reusable experience.}
\label{fig:memorizer-architecture}
\end{wrapfigure}

EMem is an external memory system that continually organizes and updates world state during embodied interaction. Figure~\ref{fig:memorizer-architecture} provides an overview of the EMem architecture.

\Needspace{4\baselineskip}
\subsection{Three Complementary Memories}

\mypar{Spatial memory.}
Spatial memory maintains an entity graph that tracks the locations and states of objects. New observations and feedback update the corresponding records so that the graph reflects the latest known state of each object. For example, when a book is moved from the bed to the desk, spatial memory replaces the outdated relation to the bed and returns the desk as the book's current location.

\mypar{Event memory.}
Event memory preserves actions, environment feedback, and human corrections in temporal order. It can also summarize a pattern across several corrections and retrieve that pattern when the agent encounters a related new object. For example, when the agent tries to open the top drawer and learns that it is locked, event memory preserves the attempt and its outcome so that the agent can avoid repeating the same action later.

\mypar{Scene memory.}
Scene memory associates objects and interactions with their scenes, preventing confusion between same-named entities in different rooms. For example, when mugs appear in both the kitchen and the bedroom, scene memory binds each mug and its interactions to the corresponding room, allowing the correct instance to be retrieved.

\subsection{Operating the Memory Loop}
\label{sec:sft}

EMem provides write and query interfaces for each memory. During interaction, the model stores visual observations, environmental changes, action outcomes, and corrections. For a target task, it queries one or more memories and combines the returned information with the current observation to generate a complete action sequence. Execution produces the next observation and feedback, allowing memory to be updated as interaction continues. The complete tool interface and retrieval procedure are provided in Appendix~\ref{app:memorizer-details}.

We use supervised fine-tuning to teach three decisions: what to write from the current observation, which memories to query for the target task, and how to generate an action sequence from the retrieved information. Training trajectories are generated from the ProcTHOR training split, with a teacher model providing supervision. Fine-tuning Qwen3-VL-8B produces \emph{EMem-8B}; data construction and full training details are provided in Appendix~\ref{app:memorizer-details}.

% End the method figure's wrapping before the next section, even across a page break.
\WFclear
\section{Experiments}
\label{sec:experiments}

\subsection{Experimental Setup}

\mypar{Baselines.}
Our general-purpose open-source baselines include Qwen3-VL~\citep{bai2025qwen3vl}, InternVL3~\citep{zhu2025internvl3}, Ovis2~\citep{lu2025ovis25}, Qwen3.6~\citep{qwen2026qwen36}, and Mistral-Small-3.1~\citep{mistral2025small31}. The embodied MLLMs include RynnBrain~\citep{dang2026rynnbrain}, Cambrian-S~\citep{yang2025cambrians}, RoboBrain2.5~\citep{tan2026robobrain25}, MiMo-Embodied~\citep{hao2025mimoembodied}, and the Robotics-ER 1.5~\citep{geminirobotics2025}. Our general-purpose proprietary baselines include GPT-5.4 and GPT-5.4-mini~\citep{singh2025gpt5}, Gemini-2.5-Pro~\citep{geminiteam2025gemini25}, and Gemini-3-Flash~\citep{google2025gemini3flash}. To fit episodes within the context, the \texttt{full\_context} condition retains the complete trajectory in text form and only the final visual observation. For the memory-system comparison, we fix GPT-5.4-mini as the backbone and evaluate MIRIX~\citep{wang2025mirix}, MemVerse~\citep{liu2025memverse}, TeleMem~\citep{chen2025telemem}, MMA~\citep{lu2026mma}, and EMem.

\subsection{Main Results}

\begin{table}[H]
\centering
\setlength{\tabcolsep}{1.5pt}
\renewcommand{\arraystretch}{0.98}
{\fontsize{9}{9.5}\selectfont
\begin{tabular}{@{}>{\raggedright\arraybackslash}p{100pt}@{\hspace{0.8pt}}*{10}{>{\centering\arraybackslash}p{\dimexpr(\linewidth-108.8pt)/10\relax}@{\hspace{0.8pt}}}@{}}
\toprule
\multirow[c]{2}{100pt}[-3pt]{\centering \textbf{Model / system}} & \multicolumn{2}{c}{\textbf{Average}} & \multicolumn{2}{c}{\textbf{Passive}} & \multicolumn{2}{c}{\textbf{Dynamic}} & \multicolumn{2}{c}{\textbf{Interaction}} & \multicolumn{2}{c}{\textbf{Exp. Gen.}} \\
\cmidrule(lr){2-3}\cmidrule(lr){4-5}\cmidrule(lr){6-7}\cmidrule(lr){8-9}\cmidrule(lr){10-11}
& SR$\uparrow$ & ERR$\downarrow$ & SR$\uparrow$ & ERR$\downarrow$ & SR$\uparrow$ & ERR$\downarrow$ & SR$\uparrow$ & ERR$\downarrow$ & SR$\uparrow$ & ERR$\downarrow$ \\
\midrule
\multicolumn{11}{l}{\textit{Open-source MLLMs}} \\
Qwen3-VL-4B-Ins & 30.1 & 37.4 & 46.4 & 36.0 & 37.3 & 45.4 & 20.9 & 22.8 & 15.8 & 45.3 \\
Qwen3-VL-8B-Ins & 25.4 & 41.1 & 41.4 & 45.0 & 26.6 & 58.9 & 11.8 & 22.1 & 21.7 & 38.4 \\
Qwen3-VL-32B-Ins & 43.9 & 38.3 & 47.1 & 39.5 & 20.3 & 70.6 & 45.3 & 24.7 & 63.1 & 18.2 \\
InternVL3-8B & 15.5 & 48.8 & 20.5 & 62.4 & 31.8 & 55.9 & 5.3 & \bestgroup{10.3} & 4.4 & 66.5 \\
InternVL3-38B & 31.8 & 47.4 & 47.9 & 33.7 & 17.0 & 64.5 & 29.3 & 29.3 & 33.0 & 62.1 \\
Ovis2-16B & 29.0 & 48.0 & 44.5 & 43.8 & 22.7 & 74.2 & 30.0 & 36.1 & 18.7 & 37.9 \\
Qwen3.6-27B & \bestgroup{58.2} & \bestgroup{25.1} & \bestgroup{61.0} & \bestgroup{19.8} & \bestgroup{49.5} & \bestgroup{39.1} & \bestgroup{54.4} & 17.9 & \bestgroup{68.0} & 23.7 \\
Mistral-Small-3.1-24B & 40.9 & 35.8 & 58.5 & 28.6 & 25.7 & 67.9 & 11.8 & 30.8 & 67.5 & \bestgroup{15.8} \\
\midrule
\multicolumn{11}{l}{\textit{Embodied MLLMs}} \\
RynnBrain-8B & 3.6 & 37.5 & 5.3 & 38.9 & 4.0 & 46.6 & 2.7 & 28.1 & 2.5 & 36.5 \\
Cambrian-S-7B & 0.8 & 61.4 & 0.8 & 83.3 & 0.9 & 67.2 & 1.1 & 36.5 & 0.5 & 58.6 \\
RoboBrain2.5-8B-NV & 20.0 & 42.3 & 34.1 & 57.4 & 29.4 & 68.4 & 16.4 & 25.1 & 0.0 & \bestgroup{18.2} \\
MiMo-Embodied-7B & 38.8 & 30.8 & 50.2 & 21.5 & 35.6 & 43.4 & 22.8 & 23.2 & 46.8 & 35.0 \\
Robotics-ER 1.5$^{*}$ & \bestgroup{52.4} & \bestgroup{25.2} & \bestgroup{60.2} & \bestgroup{19.4} & \bestgroup{47.6} & \bestgroup{39.1} & \bestgroup{44.4} & \bestgroup{18.5} & \bestgroup{57.1} & 23.8 \\
\midrule
\multicolumn{11}{l}{\textit{EMem with open-source backbones}} \\
\rowcolor{OursRow}[0.4pt][0.4pt]
Qwen3-VL-8B-Ins & \gaincell{36.2}{+10.8} & \gaincell{28.3}{-12.8} & \gaincell{44.1}{+2.7} & \gaincell{39.0}{-6.0} & \gaincell{57.1}{+30.5} & \gaincell{26.2}{-32.7} & \gaincell{16.0}{+4.2} & \gaincell{17.5}{-4.6} & \gaincell{27.6}{+5.9} & \gaincell{30.5}{-7.9} \\
\rowcolor{OursRow}[0.4pt][0.4pt]
Qwen3-VL-32B-Ins & \gaincell{\bestgroup{58.0}}{+14.1} & \gaincell{13.9}{-24.4} & \gaincell{\bestgroup{68.9}}{+21.8} & \gaincell{18.8}{-20.7} & \gaincell{55.6}{+35.3} & \gaincell{24.2}{-46.4} & \gaincell{47.9}{+2.6} & \gaincell{\bestoverall{3.0}}{-21.7} & \gaincell{59.6}{-3.5} & \gaincell{\bestoverall{9.4}}{-8.8} \\
\rowcolor{OursRow}[0.4pt][0.4pt]
Mistral-Small-3.1-24B & \gaincell{57.3}{+16.4} & \gaincell{\bestoverall{12.4}}{-23.4} & \gaincell{52.1}{-6.4} & \gaincell{\bestoverall{17.4}}{-11.2} & \gaincell{53.0}{+27.3} & \gaincell{\bestoverall{15.4}}{-52.5} & \gaincell{\bestgroup{54.8}}{+43.0} & \gaincell{3.8}{-27.0} & \gaincell{\bestgroup{69.5}}{+2.0} & \gaincell{12.8}{-3.0} \\
\rowcolor{OursRow}[0.4pt][0.4pt]
EMem-8B & \gaincell{46.0}{+20.6} & \gaincell{23.5}{-17.6} & \gaincell{55.0}{+13.6} & \gaincell{30.6}{-14.4} & \gaincell{\bestgroup{66.9}}{+40.3} & \gaincell{21.1}{-37.8} & \gaincell{23.6}{+11.8} & \gaincell{15.2}{-6.9} & \gaincell{38.4}{+16.7} & \gaincell{27.1}{-11.3} \\
\midrule
\multicolumn{11}{l}{\textit{Proprietary MLLMs}} \\
GPT-5.4-mini & 35.3 & 38.8 & 55.1 & 32.3 & 25.8 & 63.6 & 19.0 & 17.5 & 41.4 & 41.9 \\
GPT-5.4 & 56.8 & 27.1 & 60.4 & 21.9 & 46.7 & 42.5 & 51.0 & 20.2 & 69.0 & 23.7 \\
Gemini-2.5-Pro & 60.8 & 22.4 & 63.5 & 19.4 & 51.6 & 38.6 & 55.1 & 16.4 & 72.9 & \bestgroup{15.3} \\
Gemini-3-Flash & \bestoverall{64.2} & \bestgroup{21.3} & \bestgroup{67.0} & \bestgroup{18.5} & \bestgroup{53.9} & \bestgroup{34.9} & \bestoverall{60.5} & \bestgroup{14.1} & \bestoverall{75.4} & 17.7 \\
\midrule
\multicolumn{11}{l}{\textit{Multimodal memory systems with GPT-5.4-mini}} \\
\texttt{full\_context} & 35.3 & 38.8 & 55.1 & 32.3 & 25.8 & 63.6 & 19.0 & 17.5 & 41.4 & 41.9 \\
MIRIX & 28.2 & 46.7 & 49.2 & 39.7 & 20.0 & 77.9 & 11.0 & 20.5 & 32.5 & 48.8 \\
MemVerse & 28.9 & 45.1 & 49.6 & 39.6 & 18.1 & 70.1 & 14.5 & 24.0 & 33.5 & 46.8 \\
TeleMem & 37.8 & 35.7 & 49.9 & 30.3 & \bestoverall{72.0} & \bestgroup{17.2} & 4.2 & 37.3 & 25.1 & 58.1 \\
MMA & 23.4 & 47.2 & 43.8 & 38.7 & 15.5 & 76.1 & 10.3 & 22.8 & 24.1 & 51.2 \\
\rowcolor{OursRow}[0.4pt][0.4pt]
EMem & \gaincell{\bestgroup{58.9}}{+23.6} & \gaincell{\bestgroup{24.3}}{-14.5} & \gaincell{\bestoverall{71.2}}{+16.1} & \gaincell{\bestgroup{18.2}}{-14.1} & \gaincell{58.1}{+32.3} & \gaincell{35.0}{-28.6} & \gaincell{\bestgroup{44.1}}{+25.1} & \gaincell{\bestgroup{17.1}}{-0.4} & \gaincell{\bestgroup{62.1}}{+20.7} & \gaincell{\bestgroup{27.1}}{-14.8} \\
\bottomrule
\end{tabular}}
\caption{Full results on EMem-Bench (\%). Average denotes the mean across the four task families. Parenthesized gray values show absolute changes from the corresponding full-context backbone. Underlining marks the best result within each setting, and bold marks the best result overall. Robotics-ER 1.5$^{*}$ is evaluated on a 10\% subset because of API service instability.}
\label{tab:leaderboard}
\end{table}

\mypar{Overall performance.}
As shown in Table~\ref{tab:leaderboard}, current models still struggle on EMem-Bench: Gemini-3-Flash, the strongest full-context model, reaches only 64.2\% average SR. Most models exhibit markedly uneven profiles across the four task families; for example, Qwen3-VL-32B obtains 20.3\% SR on Dynamic Tracking but 63.1\% on Experience Generalization. Stronger models, in contrast, perform well across all four families rather than excelling on only one, indicating that these abilities form complementary components of long-horizon embodied interaction.

\mypar{Embodied reasoning models.}
Specialization for embodied or spatial reasoning does not consistently translate into effective memory use over long-horizon interaction. RynnBrain-8B and RoboBrain2.5-8B-NV obtain only 3.6\% and 20.0\% average SR, both below the Qwen3-VL-8B-Instruct at 25.4\%. Cambrian-S-7B reaches only 0.8\%. MiMo-Embodied-7B reaches 38.8\%, and Robotics-ER 1.5 reaches 52.4\%, yet neither surpasses the strongest general-purpose open model. These results indicate that specialization for spatial perception, embodied grounding, or robot planning alone does not equip a model to preserve, revise, and act on world state throughout extended interaction.

\mypar{Effectiveness of EMem.}
EMem achieves the best overall result, reaching 58.9\% average SR, 21.1 points above the strongest alternative, TeleMem, while reducing ERR by 11.4 points. Its gains are not limited to one backbone: EMem improves average SR by 10.8, 14.1, and 16.4 points on the three open-source backbones, respectively, and by 23.6 points on the proprietary GPT-5.4-mini. Training Qwen3-VL-8B to write, retrieve, and use the memories provides a further 9.8-point gain over applying EMem to the frozen backbone, showing that learning the memory loop further strengthens the model's use of external memory.

\subsection{Does More Visual History Help?}
\label{sec:visual-history-expansion}

Can EMem-Bench be solved by simply appending more historical observations to the model context? We test this with GPT-5.4-mini on a fixed 800-episode subset, with 200 episodes per task family. Both conditions retain the complete textual history and the current RGB observation. The expanded condition additionally supplies as many historical RGB observations as the remaining context budget permits, preserving their temporal order. The target tasks and action budget remain unchanged.

\begin{table}[H]
\centering
{\small
\setlength{\tabcolsep}{4pt}
\renewcommand{\arraystretch}{1.1}
\begin{tabular*}{\linewidth}{@{\extracolsep{\fill}}lrrr@{}}
\toprule
\textbf{Input setting} & \textbf{Avg SR (\%)} & \textbf{Input tokens} & \textbf{s/step} \\
\midrule
Standard \texttt{full\_context} & 36.6 & 41{,}770 & 4.1 \\
+ maximum historical RGB & 26.2 & 206{,}489 & 39.1 \\
\bottomrule
\end{tabular*}}
\caption{Expanding historical visual context on the same 800 episodes with GPT-5.4-mini. Average SR equally weights the four task families. Input tokens are per-episode means; model-step latency excludes simulator rendering.}
\label{tab:visual-history-expansion}
\end{table}

Table~\ref{tab:visual-history-expansion} shows that appending historical images reduces average SR from 36.6\% to 26.2\%, a paired decrease of 10.4 points, while increasing input tokens by $4.9\times$ and model-step latency by $9.5\times$. Performance declines in every history-length quartile. These results demonstrate that \textbf{EMem-Bench cannot be solved by simply enlarging the raw multimodal context}. Effective memory requires recovering and using task-relevant evidence across a long interaction history; simply accumulating more observations does not resolve this challenge. Appendix~\ref{app:historical-visual-input} provides the image-selection procedure and detailed task-family results.

\subsection{Ablation Study}

Table~\ref{tab:memory-ablation} uses GPT-5.4-mini as the fixed backbone and disables one memory at a time while keeping the other two active.

\begin{table}[H]
\setlength{\tabcolsep}{1.5pt}
\renewcommand{\arraystretch}{1.1}
\centering
{\small
\setlength{\tabcolsep}{0.5pt}
\begin{tabular*}{\columnwidth}{@{\extracolsep{\fill}}lrrrrr@{}}
\toprule
\textbf{Setting} & \textbf{\mbox{Avg}} & \textbf{\mbox{Passive}} & \textbf{\mbox{Dynamic}} & \textbf{\mbox{Interaction}} & \textbf{\mbox{Exp. Gen.}} \\
\midrule
Full system & 58.9 & 71.2 & 58.1 & 44.1 & 62.1 \\
w/o spatial & 47.9 & 59.1 & 46.1 & 36.1 & 50.3 \\
w/o event & 46.9 & 60.1 & 54.6 & 27.4 & 45.3 \\
w/o scene & 47.3 & 51.7 & 49.5 & 31.9 & 56.2 \\
\bottomrule
\end{tabular*}}
\caption{Ablation of the three memories in EMem.}
\label{tab:memory-ablation}
\end{table}

Removing any memory reduces average SR, showing that all three contribute substantially to the full system. The task-wise drops reveal distinct roles: Dynamic Tracking depends most on spatial memory; Interaction Failure and Experience Generalization depend most on event memory; and Passive Observation depends most on scene memory.

\subsection{Cross-Benchmark Generalization}

The EB-ALF suite of EmbodiedBench tests transfer to general embodied execution; MMMU-Pro \citep{yue2024mmmupro} evaluates general multimodal understanding; BLINK \citep{fu2024blink} tests fine-grained visual perception; and HallusionBench \citep{guan2024hallusionbench} evaluates robustness to visual illusion and language hallucination.

\begin{table}[H]
\centering
\setlength{\tabcolsep}{1.5pt}
\renewcommand{\arraystretch}{1.1}
{\small
\setlength{\tabcolsep}{1pt}
\begin{tabular*}{\columnwidth}{@{\extracolsep{\fill}}lrrrr@{}}
\toprule
\textbf{Model} & \textbf{\mbox{EB-ALF}} & \textbf{\mbox{MMMU-Pro}} & \textbf{\mbox{BLINK}} & \textbf{\mbox{HallusionBench}} \\
\midrule
Qwen3-VL-8B & 18.3 & 44.6 & 55.6 & 66.1 \\
EMem-8B & 23.3 & 45.1 & 56.9 & 65.8 \\
\bottomrule
\end{tabular*}}
\caption{Cross-benchmark generalization results.}
\label{tab:external-generalization}
\end{table}

Table~\ref{tab:external-generalization} compares the base Qwen3-VL-8B with EMem-8B across all four benchmarks. EMem-8B improves average SR on EB-ALF by 5.0 points and macro accuracy on BLINK by 1.3 points. It also gains 0.5 points on MMMU-Pro, while remaining comparable on HallusionBench with a 0.4-point decrease. These results show that memory-operation training transfers most strongly to interactive embodied execution, also improves fine-grained visual perception, and preserves general multimodal perception and reasoning.

\Needspace{4\baselineskip}
\subsection{Failure Analysis}

\begin{wrapfigure}{R}{0.45\textwidth}
\vspace{0pt}
\centering
\includegraphics[width=\linewidth]{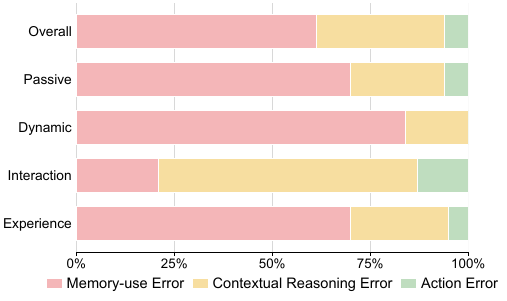}
\caption{Error distribution of GPT-5.4-mini failures.}
\label{fig:failure-analysis}
\end{wrapfigure}

We analyze 400 failed GPT-5.4-mini episodes under the full-context setting, sampling 100 from each task family. A \emph{memory-use error} occurs when the agent's behavior conflicts with earlier evidence, such as relying on an outdated location or disregarding a learned correction. A \emph{contextual reasoning error} occurs when the relevant history is available but but the agent fails to interpret or use it. For example, the agent may see that a KeyChain and a CreditCard were corrected to the same location but fail to infer that the same rule applies to a Watch. An \emph{action error} occurs when the target object and relation are correct but the action sequence goes wrong. For example, the agent may select correct object and receptacle but fail to open the receptacle before placing the object.

Figure~\ref{fig:failure-analysis} reveals distinct failure patterns across task families. Memory-use errors dominate Passive Observation, Dynamic Tracking, and Experience Generalization. Their concentration in Dynamic Tracking is revealing: the difficulty is not retaining more history, but replacing outdated state with the latest evidence. Interaction Failure shows a different pattern. Most failures arise from contextual reasoning, indicating that avoiding a previously failed interaction does not necessarily lead to a revised decision. Action errors remain uncommon across all four families, suggesting that the primary bottleneck lies before action generation. Overall, current agents struggle both to maintain the right state from experience and to reason over that state in the current task.

\section{Conclusion}

We introduce EMem-Bench, which evaluates fine-grained visual memory, dynamic world-state tracking, recording world state through interaction, and experience generalization through four executable task families. We further present Embodied-Memorizer, which organizes multimodal experience through spatial, event, and scene memory and trains an agent to write, retrieve, and use these memories during interaction. Experiments show that current MLLMs still struggle to maintain and reuse world state over long-term interaction, while EMem improves memory-guided behavior.

% Preserve the source manuscript's explicitly included bibliography entries.
\nocite{openbmb2026minicpmv,gemmateam2026gemma4,glmteam2025glm45v}
\FloatBarrier
\bibliographystyle{assets/acl_natbib}
\bibliography{paper}

\clearpage
\beginappendix
\section{Extended Related Work}
\label{app:extended-related-work}

\mypar{History-aware embodied benchmarks.}
Recent embodied benchmarks increasingly incorporate historical observations, prior interactions, or stateful trajectories into specific task settings. STARBench couples temporal recall with spatial action in open-world object retrieval~\citep{chen2025star}. RoboMME evaluates temporal, spatial, object, and procedural memory in long-horizon manipulation, while RoboMemArena extends robotic evaluation to longer trajectories, composed subtasks, and paired real-world tasks~\citep{dai2026robomme,lei2026robomemarena}. RoboMME-Interference further measures how retrieval degrades as unrelated sessions accumulate between a relevant demonstration and the query~\citep{rathi2026interference}. eMEM-Bench organizes ProcTHOR probes around source monitoring, interference, and context-dependent retrieval~\citep{rasheed2026emem}. Sequential embodied question answering also shows that carrying an occupancy map or unstructured history across questions does not preserve the visual-semantic state needed for later decisions~\citep{cai2026sequentialeqa}. Together with FindingDory, LMEE-Bench, SpaMEM, WorldLines, and WorldMemArena~\citep{yadav2025findingdory,wang2026lmee,liao2026spamem,zhang2026worldlines,liu2026worldmemarena}, these benchmarks test historical information within navigation, manipulation, question answering, or state reconstruction. EMem-Bench instead organizes complementary memory demands within a single interactive evaluation.

\mypar{Related long-horizon settings.}
3DMem-Bench combines embodied tasks, question answering, and captioning to study long-term spatial-temporal reasoning, while 3DLLM-Mem selectively fuses features from past observations for reasoning and action~\citep{hu20253dllmmemlongtermspatialtemporalmemory}. GOAT-Bench studies lifelong navigation through sequences of category, language, and image goals and analyzes the role of scene representations across successive targets~\citep{khanna2024goatbenchbenchmarkmultimodallifelong}. OST-Bench evaluates online spatio-temporal understanding from observations gathered during active exploration, including how performance changes as the exploration horizon grows~\citep{lin2025ostbenchevaluatingcapabilitiesmllms}. PersONAL evaluates navigation and object grounding from user-specific object--owner associations~\citep{ziliotto2025personalcomprehensivebenchmarkpersonalized}. Together, these benchmarks introduce long-range dependence through spatial-temporal reasoning, lifelong navigation, scene understanding, or personalized grounding. EMem-Bench adopts a different organizing view: whether information acquired during interaction is maintained, updated, and reused when later environment actions depend on it, with separate executable task families for fine-grained visual recall, dynamic state revision, interaction-revealed state, and experience transfer.

\mypar{Long-horizon multimodal memory evaluation.}
Long-form visual and conversational benchmarks expose memory operations that are difficult to study through a single embodied rollout. EgoMemReason evaluates entity-state evolution, temporally separated events, and recurring behavior patterns over week-long egocentric video~\citep{wang2026egomemreason}. Mem-Gallery studies multimodal memory extraction, reasoning, and knowledge management across multi-session conversations~\citep{bei2026memgallery}. Together with EgoSchema, OpenEQA, LoCoMo, and LongMemEval~\citep{mangalam2023egoschema,majumdar2024openeqa,maharana2024locomo,wu2025longmemeval}, these works provide controlled tests of long-range evidence integration. Their outputs are primarily answers or reconstructed knowledge, whereas embodied execution additionally requires retrieved evidence to determine actions and change the environment correctly.

\mypar{Memory architectures for embodied control.}
Recent methods differ in what they store and how memory enters the control loop. Memo learns when to summarize and retrieve history for long-horizon embodied policies~\citep{gupta2025memo}. MemoryVLA combines working memory with a bank of perceptual detail and semantic content, while MEM represents short-range history with video and long-range task progress with language~\citep{shi2025memoryvla,torne2026mem}. STaR builds task-agnostic multimodal memory and retrieves a compact task-conditioned subset; HIMM separates episodic recall from reusable semantic rules for exploration and question answering~\citep{yuan2026star,li2026himm}. MemCompiler conditions memory selection on the agent's current execution state, addressing the mismatch between static retrieved context and a changing task state~\citep{ding2026memcompiler}. Affordance RAG further grounds retrieved visual memory in executable mobile-manipulation choices~\citep{korekata2025affordancerag}. These systems motivate structured and selective memory, while EMem focuses on maintaining scene, spatial, and event records that jointly support the four challenges measured by EMem-Bench.

\mypar{Learning from interaction outcomes.}
Interaction creates information that is absent from passive observation. Memory-augmented manipulation methods use previous successes, failures, and corrections to adapt later decisions. In particular, MEMO aggregates local language corrections and successful executions into reusable guidance and skill templates for new tasks~\citep{christie2026memo}. This direction is closely related to the Interaction Failure and Experience Generalization families in EMem-Bench: the former tests whether an outcome updates the represented world state, and the latter tests whether recurring corrections yield a rule that transfers to an unseen object.

\section{Embodied-Memorizer Details}
\label{app:memorizer-details}

This section expands the three-memory design in Section~\ref{sec:method} and describes the implementation used in our experiments. EMem is deliberately modular: a policy decides what to write, what to retrieve, and how to act, while the memory backend performs deterministic record updates and searches. All memory is episode-local and is initialized empty for each trajectory.

\subsection{Memory Representation}

At interaction step $t$, EMem maintains
\begin{equation}
\mathcal{M}_t=(\mathcal{G}_t,\mathcal{E}_t,\mathcal{C}_t),
\label{eq:emem-state-simple}
\end{equation}
where $\mathcal{G}_t$, $\mathcal{E}_t$, and $\mathcal{C}_t$ denote Spatial, Event, and Scene Memory, respectively. Each record contains an identifier, a write step, a scene scope, and links to the model-visible observation or interaction from which it was written. These shared fields let the three stores refer to the same experience without introducing a separate memory type.

\mypar{Spatial Memory.}
$\mathcal{G}_t$ is a scene-scoped entity graph. A node stores an entity name and type, its observable properties, and references to related event and scene records. Edges store relations such as \texttt{in}, \texttt{on}, \texttt{held\_by}, and \texttt{seen\_in}. Scene scope is part of the entity key, so similarly named objects in different rooms remain separate.

Spatial Memory represents the latest known state rather than an unedited history of state descriptions. Relations are grouped into families that cannot hold simultaneously. For example, \texttt{in}, \texttt{on}, and \texttt{held\_by} belong to the physical-location family. If a new record supports a relation $r$ for entity $v$, EMem updates that family as
\begin{equation}
R_t(v)=\bigl(R_{t-1}(v)\setminus F(r)\bigr)\cup\{r\},
\label{eq:emem-relation-replacement}
\end{equation}
where $F(r)$ is the set of active relations that conflict with $r$. Other relation families and compatible properties are preserved. Thus, observing a book on a desk replaces an earlier location on a bed without deleting unrelated information about the book.

\mypar{Event Memory.}
$\mathcal{E}_t$ is an append-only sequence of action outcomes. A raw event stores the action, success flag, environment feedback, a short summary, its scene scope, and supporting identifiers. Both successful and failed interactions are retained. Consequently, feedback such as a locked drawer remains available even when that state was not visually observable.

Event Memory also stores a compact aggregate when several related corrections express the same regularity. The aggregate consists of a short natural-language rule and the identifiers of the raw events that support it. We form an aggregate after at least two related correction events; an explicit persistent failure can be retained after one event. Aggregation is part of Event Memory, rather than a separate ``experience'' store, and \texttt{query\_event} can request either raw events or aggregate rules. This matches the design in the main paper: reusable experience is a summary over interaction events.

\mypar{Scene Memory.}
$\mathcal{C}_t$ stores selected visual snapshots. Each record contains a model-generated caption, the entities identified by the model in that image, the scene scope, and a reference to the original RGB observation. The image reference is attached by the harness, so the model does not construct file paths. A practical admission gate suppresses near-duplicate snapshots while retaining views that add visual content, newly observed entities, or useful instance context. The entity list is extracted from the current RGB observation; it is not a simulator-provided inventory of all objects in the environment. Scene Memory therefore preserves visual detail, while Spatial Memory maintains the corresponding structured state.

Table~\ref{tab:emem-tools} lists the complete model-visible memory interface. The tools are generic across instructions and task families; there is no task-family-specific memory or query operation.

\begin{table}[t]
\centering
\small
\setlength{\tabcolsep}{4.2pt}
\begin{tabular*}{\textwidth}{@{\extracolsep{\fill}}lllp{0.48\textwidth}}
\toprule
\textbf{Tool} & \textbf{Operation} & \textbf{Store} & \textbf{Principal fields or returned content} \\
\midrule
\texttt{update\_object} & Write/update & Spatial & Entity name and type, scene scope, relations, and observable properties \\
\texttt{remember\_event} & Append & Event & Action, outcome, feedback, and concise event summary \\
\texttt{remember\_scene} & Append & Scene & Caption, entities identified in the current image, and harness-bound image reference \\
\texttt{query\_spatial} & Retrieve & Spatial & Current entity properties and relations with supporting record identifiers \\
\texttt{query\_event} & Retrieve & Event & Raw action outcomes or evidence-backed aggregate rules \\
\texttt{query\_scene} & Retrieve & Scene & Relevant captions, scene scopes, entity indices, and linked images \\
\bottomrule
\end{tabular*}
\caption{Model-visible EMem interface. Write and query operations align directly with the three memories described in the main paper.}
\label{tab:emem-tools}
\end{table}

\subsection{Writing Memory from Interaction History}

EMem ingests an interaction history chronologically. At history step $i$, the writer receives only the current model-visible RGB observation $o_i^{\mathrm h}$, the preceding action $a_{i-1}^{\mathrm h}$ and feedback $f_{i-1}^{\mathrm h}$ when available, the scene scope, and the three write-tool schemas. The future target instruction is not available during this stage. The update is
\begin{equation}
\mathcal{M}_i=\operatorname{Apply}\!\left(
\mathcal{M}_{i-1},
W_{\theta}(o_i^{\mathrm h},a_{i-1}^{\mathrm h},f_{i-1}^{\mathrm h})
\right),
\label{eq:emem-write-simple}
\end{equation}
where $W_{\theta}$ is the model policy and \textsc{Apply} executes only schema-valid write calls. The policy may call any subset of \texttt{update\_object}, \texttt{remember\_event}, and \texttt{remember\_scene}. The backend then performs the corresponding relation replacement, event append/consolidation, or scene admission operation. Record identifiers link an event or scene to the entities created from the same model-visible evidence.

The same write path is used after task-time actions. Once an action is executed, its new observation and environment feedback are passed to the writer, allowing later decisions to use the resulting state instead of a stale pre-action record.

\subsection{Task-Conditioned Retrieval and Execution}

When the target instruction becomes available, the policy enters a memory phase. It selects one or more of \texttt{query\_spatial}, \texttt{query\_event}, and \texttt{query\_scene} according to the information needed by the instruction. Query arguments specify a natural-language information need and, where applicable, scene scope, temporal policy, event granularity, and a maximum number of records. The query must not presuppose its answer. Experience Generalization uses the same \texttt{query\_event} interface as any other instruction and does not invoke a benchmark-specific tool.

Each query searches its associated store with the same deterministic pipeline. EMem first applies available structured constraints, including memory type, scene scope, entity name, relation, action, or event granularity. It then prioritizes exact structured matches, followed by lexical matches over record text. Semantic similarity is used only as a fallback when the earlier stages do not fill the requested record budget. The primary selected set is capped at 12 records; returned records also carry identifiers for their direct support when available. For a spatial query with the \texttt{latest} policy, newer records are preferred; persistent event feedback and aggregate rules are not discarded merely because they are old.

The retrieved records are inserted into the action context together with the instruction, current RGB observation, and current legal-action list. The policy then emits one ordered \texttt{action\_sequence}. Before execution, every primitive is checked against the refreshed legal actions and its target must be grounded by the current interface or retrieved records. The environment executes accepted actions in order. A failed or unavailable action, or a sufficiently large change in the memory-backed state, interrupts the remaining sequence and starts a new memory--action turn. Algorithm~\ref{alg:emem-loop-simple} summarizes this procedure.

\begin{algorithm}[tb]
\caption{EMem write--retrieve--act loop}
\label{alg:emem-loop-simple}
\textbf{Input}: history $\mathcal H$; target task $\mathcal T$; action budget $B$\\
\textbf{Output}: executed trajectory $A$
\begin{algorithmic}[1]
\STATE Initialize $\mathcal M_0\leftarrow\emptyset$
\FOR{each step $(o_i^{\mathrm h},a_{i-1}^{\mathrm h},f_{i-1}^{\mathrm h})$ in $\mathcal H$}
    \STATE $U_i\leftarrow W_\theta(o_i^{\mathrm h},a_{i-1}^{\mathrm h},f_{i-1}^{\mathrm h})$
    \STATE Apply schema-valid calls in $U_i$ to $\mathcal M_{i-1}$
\ENDFOR
\STATE $A\leftarrow[\,]$
\WHILE{the task is unfinished and $|A|<B$}
    \STATE Observe current RGB $o_t$ and legal actions $\mathcal A_t$
    \STATE $Q_t\leftarrow\textsc{SelectQueries}(\mathcal T,o_t)$
    \STATE $R_t\leftarrow\textsc{Retrieve}(\mathcal M_t,Q_t)$
    \STATE $P_t\leftarrow\textsc{Plan}(\mathcal T,o_t,\mathcal A_t,R_t)$
    \FOR{each action $a$ in the ordered plan $P_t$}
        \IF{$a\notin\mathcal A_t$}
            \STATE break and begin a new memory--action turn
        \ENDIF
        \STATE $(o_{t+1},f_t)\leftarrow\textsc{Execute}(a)$; append $a$ to $A$
        \STATE Update $\mathcal M_t$ from $(o_{t+1},a,f_t)$
        \IF{$f_t$ is a failure or the represented state changes materially}
            \STATE break and begin a new memory--action turn
        \ENDIF
    \ENDFOR
\ENDWHILE
\STATE \textbf{return} $A$
\end{algorithmic}
\end{algorithm}

The runtime settings are fixed across benchmark families: the primitive-action budget is 8, the primary retrieval budget is at most 12 records per call, near-duplicate scenes use a cosine-similarity threshold of 0.995, and evaluation uses deterministic decoding. These values are not selected using benchmark-family labels.

\subsection{Prompt Templates}

Figures~\ref{fig:context-ingestion-prompt} and~\ref{fig:task-action-prompt} show the prompts used for context ingestion and task execution. Bracketed fields are populated at runtime with model-visible observations, legal actions, declared tool schemas, or the results of executed memory calls. The controller validates the returned JSON against the active phase: write tools are available only during ingestion, memory queries only during the memory phase, and robot actions only after retrieval.

\newsavebox{\promptbodybox}

\begin{figure}[!htbp]
\centering
\begingroup
\setlength{\fboxsep}{5.5pt}
\setlength{\fboxrule}{0.8pt}
\noindent\colorbox{black}{%
  \parbox{\dimexpr\linewidth-2\fboxsep\relax}{%
    \color{white}\bfseries Context-ingestion prompt template.}}
\par\vspace{-0.8pt}
\begin{lrbox}{\promptbodybox}
\begin{minipage}{\dimexpr\linewidth-2\fboxsep-2\fboxrule\relax}
\begin{lstlisting}[
  basicstyle=\fontsize{7.4}{8.5}\selectfont\ttfamily,
  numbers=none,
  xleftmargin=0pt,
  frame=none,
  aboveskip=1pt,
  belowskip=1pt,
  breaklines=true,
  breakatwhitespace=true,
  columns=fullflexible,
  keepspaces=true,
  showstringspaces=false
]
[SYSTEM]
You are an expert embodied-memory agent controlling a household robot.

A valid interaction cycle has two ordered stages:
1. context_ingestion: process model-visible observations and call EMem update tools;
2. task_execution: query relevant memories before producing an action_sequence.

Never put action_sequence in context_ingestion. Never use retrieval tools while ingesting context.
Earlier experience must enter a later task through executed memory writes, not as an unstructured transcript.

[USER]
CONTEXT INGESTION STEP. This is one step of embodied experience.
Read only this step and store durable information that may be useful later.

Current observation:
<OBSERVATION_JSON_AND_IMAGE>

Previous action and environment feedback:
<ACTION_AND_FEEDBACK>

Available update tools:
- update_object(name, node_type, relations, properties)
- remember_event(action, success, feedback, note)
- remember_scene(caption, visible_objects)

Rules:
1. Do not assume any future instruction or requested entity at this stage.
2. Write observable object locations and states with update_object.
3. Preserve exact targets and durable feedback, including locked or invalid interactions.
4. Record meaningful action outcomes with remember_event.
5. Save a scene when it supplies useful visual context or instance disambiguation.
6. Use update tools only. Retrieval tools and robot actions are forbidden.
7. Return one JSON object and no text outside it.

Return:
{
  "phase": "context_ingestion",
  "reasoning": {
    "state": "Concise current world state.",
    "memory_update": "Durable records to add or revise.",
    "memory_retrieval": "Not allowed during context ingestion."
  },
  "tool_calls": [
    {"tool": "<UPDATE_TOOL>", "args": {<SCHEMA_VALID_ARGUMENTS>}}
  ]
}
\end{lstlisting}
\end{minipage}
\end{lrbox}
\noindent\fcolorbox{black}{EMPromptBody}{\usebox{\promptbodybox}}
\endgroup
\caption{Prompt used to convert one prior interaction step into structured EMem writes.}
\label{fig:context-ingestion-prompt}
\end{figure}

\begin{figure}[!htbp]
\centering
\begingroup
\setlength{\fboxsep}{5.5pt}
\setlength{\fboxrule}{0.8pt}
\noindent\colorbox{black}{%
  \parbox{\dimexpr\linewidth-2\fboxsep\relax}{%
    \color{white}\bfseries Task-time retrieval and action prompt template.}}
\par\vspace{-0.8pt}
\begin{lrbox}{\promptbodybox}
\begin{minipage}{\dimexpr\linewidth-2\fboxsep-2\fboxrule\relax}
\begin{lstlisting}[
  basicstyle=\fontsize{7.4}{8.5}\selectfont\ttfamily,
  numbers=none,
  xleftmargin=0pt,
  frame=none,
  aboveskip=1pt,
  belowskip=1pt,
  breaklines=true,
  breakatwhitespace=true,
  columns=fullflexible,
  keepspaces=true,
  showstringspaces=false
]
[SYSTEM]
You are an expert embodied-memory agent controlling a household robot.
Each task turn has a MEMORY PHASE followed by an ACTION PHASE.
Do not emit robot actions before the selected memory tool results are returned.

[USER -- MEMORY PHASE]
Earlier interaction context has entered EMem through executed update-tool calls.

Target instruction: <TARGET_TASK>
Current observation and legal action schema: <TASK_OBSERVATION>

First decompose the instruction into:
- entity and scene anchors;
- required relation, state, affordance, or procedure;
- temporal policy: latest, persistent, episodic, or trend-sensitive;
- evidence granularity: fact, raw_event, aggregate_rule, or visual_scene;
- scope, provenance, and evidence-coverage constraints.

Select the smallest sufficient set from the three general memory queries:
- query_spatial(query, temporal_policy, max_records)
- query_event(query, granularity, temporal_policy, max_records)
- query_scene(query, scope, max_records)

Set max_records from entity ambiguity, view uncertainty, and required evidence coverage.
The query must express an information need without presupposing its result.
Do not call update tools, action_sequence, robot actions, or undeclared tool aliases.

Return:
{
  "phase": "memory",
  "reasoning": {
    "state": "What is known from the current observation.",
    "memory_update": "No updates in this phase.",
    "memory_retrieval": "Evidence required before acting."
  },
  "tool_calls": [
    {"tool": "<QUERY_TOOL>", "args": {<SCHEMA_VALID_QUERY_ARGUMENTS>}}
  ]
}

[TOOL]
Embodied-Memorizer result(s):
<RETRIEVED_RECORDS_WITH_SUPPORTING_EVIDENCE>

[USER -- ACTION PHASE]
Return exactly one action_sequence grounded in the current observation, legal actions, or retrieved memory.
Emit the shortest complete plan within the remaining budget, including all navigation and manipulation prerequisites.
Prefer the most recent well-supported state, and preserve causal ordering among navigation, opening, pickup, and placement.
Never repeat an interaction whose durable failure condition still holds.
Do not call memory tools in the action phase.

Return:
{
  "phase": "action",
  "reasoning": {
    "memory_evidence": "Evidence used by the plan.",
    "planning": "Complete memory-grounded plan.",
    "sequence_check": "Ordered actions and action count."
  },
  "tool_calls": [
    {"tool_name": "action_sequence", "arguments": {"actions": [
      {"action_type": "<TYPE>", "target_label": "<LABEL>", "target_type": "<OBJECT_TYPE>"}
    ]}}
  ]
}
\end{lstlisting}
\end{minipage}
\end{lrbox}
\noindent\fcolorbox{black}{EMPromptBody}{\usebox{\promptbodybox}}
\endgroup
\caption{Prompt used to retrieve task-relevant memory and produce an executable action sequence.}
\label{fig:task-action-prompt}
\end{figure}

\subsection{Training Procedure}

\mypar{Teacher trajectories.}
We construct a household-interaction curriculum on the ProcTHOR training split. The sampled tasks cover navigation, object search, state inspection, manipulation, and correction-bearing interactions. A teacher model completes them through the same RGB observation, memory-tool, and action interfaces used at inference. Every trajectory begins with empty memory. During history ingestion, the teacher processes one chronological step at a time without receiving the later target task, and its valid write calls are executed before the next step. After the target is presented, the teacher queries memory and generates an ordered action sequence from the returned records. Training and evaluation use disjoint scenes and trajectories.

\mypar{Supervision format.}
Each retained trajectory is serialized into three turn types: context-write turns (C), target-task memory turns (M), and action turns (A). A C target selects durable information supported by the current observation or feedback and writes it through Table~\ref{tab:emem-tools}. An M target selects one or more of the three generic query tools. The following A target receives the executed query results and produces the grounded action sequence. Examples contain only the transcript prefix available at that point. Parent trajectories are assigned to a split before prefix expansion, ensuring that all turns from one trajectory remain in the same split. Schema validation removes outputs with an invalid phase, undeclared tool, missing argument, or malformed action sequence.

\mypar{Corpus and optimization.}
The resulting corpus contains 8,852 supervision turns: 3,930 context-write turns and 2,461 paired examples for each of the memory and action phases. Table~\ref{tab:corpus} reports the trajectory-level splits. All accepted assistant responses are optimized using the standard next-token objective.

\begin{table}[!tbp]
\centering
\small
\setlength{\tabcolsep}{4.0pt}
\begin{tabular}{@{}lrrrrr@{}}
\toprule
\textbf{Split} & \textbf{Traj.} & \textbf{Turns} & \textbf{C} & \textbf{M} & \textbf{A} \\
\midrule
Train & 77 & 7,264 & 2,464 & 2,400 & 2,400 \\
Validation & 8 & 797 & 735 & 31 & 31 \\
Held-out & 8 & 791 & 731 & 30 & 30 \\
\midrule
Total & 93 & 8,852 & 3,930 & 2,461 & 2,461 \\
\bottomrule
\end{tabular}
\caption{Memory-operation training corpus. C/M/A denote context-write, target-task memory, and action turns.}
\label{tab:corpus}
\end{table}

We fine-tune Qwen3-VL-8B with LoRA~\citep{hu2022lora} for one epoch and 1,816 optimization steps. We use rank 4, alpha 8, zero dropout, a learning rate of $10^{-4}$, batch size 1, gradient accumulation 4, and a maximum sequence length of 8,192 tokens. The system messages, phase labels, tool schemas, JSON output format, and deterministic decoding used to collect supervision are retained at inference.

\section{Benchmark Details}
\label{app:benchmark-details}

This section specifies the evaluation interface used by the \texttt{full\_context} baselines in Table~\ref{tab:leaderboard} and expands the benchmark summary in Section~\ref{sec:benchmark} using the canonical Full-2554 release. We first describe exactly how the interaction history, current observation, and legal action space are exposed to a model, then report model-output trajectory lengths, benchmark composition and difficulty, object coverage, release audit results, and aggregation sensitivity. EMem-Bench contains 2,554 episodes across 1,118 globally unique scenes, 125 visible object types, 83 target object types, and 33 receptacle types. All family-level scene and category counts below are computed within that family and may overlap across families; they therefore should not be summed to recover the global unions.

\subsection{Full-Context Evaluation Protocol}

The \texttt{full\_context} condition directly provides the episode's original interaction history as model-visible textual records. For each context step, the record retains the semantic action type, target label and type, a sanitized natural-language action, success flag, environment feedback, and any model-visible observation note. Session names, descriptions, and temporal order are preserved.

Let episode $i$ contain $S_i$ history sessions, with session $j$ denoted by $\mathcal H_{i,j}$ and containing $N_{i,j}$ steps. The history supplied in the full-context condition is
\begin{equation}
\mathcal C_i^{\mathrm{FC}}
=
\bigoplus_{j=1}^{S_i}
\mathsf{San}\!\left(
\mathcal H_{i,j}[1:\min(K,N_{i,j})]
\right),
\qquad K=80,
\label{eq:full-context-serialization}
\end{equation}
where $\mathsf{San}$ performs the field filtering above and $\bigoplus$ preserves session and step order. The 80-step limit applies separately to each session. The longest session in the canonical release has 78 steps, so Equation~\ref{eq:full-context-serialization} retains the complete textual history for every Full-2554 episode.

At target-task step $t$, the actual multimodal request is
\begin{equation}
\begin{aligned}
x_{i,t}^{\mathrm{FC}}=\bigl(&\mathcal T_i,\mathcal C_i^{\mathrm{FC}},I_{i,t}^{\mathrm e},
\mathcal A(s_{i,t}),\\
&\mathcal R^{(8)}_{i,t-1}\bigr),
\end{aligned}
\label{eq:full-context-request}
\end{equation}
where $I_{i,t}^{\mathrm e}$ is the current $500\times500$ RGB observation and $\mathcal A(s_{i,t})$ is the complete legal action list at the current state. $\mathcal R^{(8)}_{i,t-1}$ retains the last eight model--environment turns, including prior actions, success flags, sanitized feedback, and compact model responses. Because evaluation permits at most eight target-task steps, this recent record contains the complete execution trajectory generated so far.

The reported full-context runs use deterministic decoding and request one next primitive action per model call. Writing the raw model response as $y_{i,t}$, the execution interface is
\begin{align}
y_{i,t}
&=M_{\theta}(x_{i,t}^{\mathrm{FC}}), \notag\\
\widetilde a_{i,t}
&=\mathsf{Parse}(y_{i,t}), \notag\\
a_{i,t}
&=\mathsf{Bind}\!\left(\widetilde a_{i,t},\mathcal A(s_{i,t})\right),
\label{eq:full-context-action}\\
(s_{i,t+1},o_{i,t+1}^{\mathrm e},f_{i,t})
&=\mathsf F(s_{i,t},a_{i,t}).
\label{eq:full-context-transition}
\end{align}
$\mathsf{Parse}$ extracts a structured action, while $\mathsf{Bind}$ resolves its action identifier or semantic target fields against the current legal action space. The evaluator records an unavailable or invalid action as a failed attempt. After every valid transition it refreshes the image and legal actions before the next request. Execution terminates when the simulator state satisfies $\mathcal T_i$, the model emits a termination action, or the eight-step budget is exhausted.

\subsection{Operational Definition of Error Recurrence Rate}

ERR measures whether an agent reenacts a mistake that the preceding interaction history makes avoidable. It is computed automatically from the bound action trace and frozen probe annotations. Each released probe specifies the memory-dependent relation and, where applicable, one or more exact error triggers used by the evaluator.

Let the executed trace for episode $i$ be $A_i=(a_{i,1},\ldots,a_{i,L_i})$. After semantic targets have been rebound to the current legal action space, the evaluator compares every successfully executed action with the probe specification. For Passive Observation, ERR is triggered when the agent, before reaching the remembered object location, successfully navigates to a different semantic receptacle type that is inconsistent with the frozen target-location annotation. For Dynamic Tracking, it is triggered when the agent, before reaching the latest post-update location, successfully navigates to a different-type superseded or otherwise incorrect location. Both families accept physical-parent aliases such as \texttt{Sink} for \texttt{SinkBasin}; a same-type alternative is left unpenalized because it may reflect instance grounding rather than recurrence of a memory error.

For Interaction Failure, ERR is triggered when the agent successfully repeats the exact action--target pair identified by the earlier failed interaction, such as opening the same known unusable container. For Experience Generalization, it is triggered when the agent successfully places the carried target object at a receptacle inconsistent with the frozen owner-habit rule and its executable target aliases. Together, these four deterministic rules identify recurrence of the family-specific avoidable error from simulator-bound semantic actions and frozen episode annotations.

For every matched action, the evaluator records
\begin{equation}
e=\langle t,\operatorname{type}(a_t),\operatorname{target}(a_t),
\operatorname{penalty}(e),\operatorname{description}(e)\rangle .
\label{eq:err-event-key}
\end{equation}
Events with the same five fields are deduplicated, preventing two equivalent detector paths from counting the same action twice. Repeating the same known mistake at a later executed step is a new recurrence because its step index differs. Let $\mathcal E_i$ be the resulting set and $E_i=|\mathcal E_i|$. Let $\mathcal Q_i$ be the probe's declared memory-integration requirements. The implementation uses
\begin{equation}
K_i=\max(1,|\mathcal Q_i|),\qquad
\mathrm{ERR}_i=\min\!\left(1,\frac{E_i}{K_i}\right).
\label{eq:operational-episode-err}
\end{equation}
Every Full-2554 episode contains one primary memory-dependent decision: the L2 probes encode it through their probe type and frozen target or interaction trigger, while each Experience Generalization probe contains one explicit integration requirement. Consequently, $K_i=1$ throughout the canonical release and $\mathrm{ERR}_i$ is the indicator that at least one qualifying recurrence occurred. Multiple repetitions remain visible in the released \texttt{triggered\_err} list but cannot make one episode contribute more than one to the family mean. Thus the reported family score is equivalently
\begin{equation}
\mathrm{ERR}_f=
\frac{|\{i\in f:\texttt{triggered\_err}_i\neq\emptyset\}|}{N_f},
\label{eq:operational-family-err}
\end{equation}
and Average ERR is the arithmetic mean of the four family ERR values, matching the equal-family aggregation used for SR.

The evaluator assigns malformed, unavailable, or simulator-failed actions to the invalid-action or RAR tallies and performs ERR matching on the bound action trace. A qualifying trigger is required for $\mathrm{ERR}_i=1$; terminal failure alone leaves $\mathrm{ERR}_i=0$, while an agent may trigger ERR, recover, and still complete the task. For example, navigating first to a Book's superseded Box location in Dynamic Tracking yields $E_i=1$ and $\mathrm{ERR}_i=1$, whereas navigating directly to its latest Desk location yields zero. Reopening the same known unusable Drawer at two different steps yields $E_i=2$ but still $\mathrm{ERR}_i=1$ after clipping. These rules keep ERR complementary to terminal-state SR: SR asks whether the final environment state is correct, while ERR asks whether the executed path reenacts an error that the episode history should have prevented.

\subsection{Model Output Trajectory Length}

For model $m$ on episode $i$, we define its output trajectory and length as
\begin{equation}
\begin{aligned}
\tau_i^{(m)}
&=\left(a_{i,1}^{(m)},\ldots,a_{i,L_i^{(m)}}^{(m)}\right),\\
L_i^{(m)}
&=\left|\tau_i^{(m)}\right|
=\texttt{actual\_steps}\\
&=\left|\texttt{model\_actions}\right|.
\end{aligned}
\label{eq:model-output-trajectory-length}
\end{equation}
Thus $L_i^{(m)}$ counts primitive action attempts recorded by the evaluator, including an invalid attempt or explicit termination action when present. We audit all 30,648 rollouts from the 12 general-purpose full-context MLLMs that completed the final 2,554-episode leaderboard. Every rollout satisfies the equality in Equation~\ref{eq:model-output-trajectory-length}; all use the shared full-context configuration specified above.

\begin{table}[!t]
\centering
\small
\setlength{\tabcolsep}{4.0pt}
\begin{tabular*}{\textwidth}{@{\extracolsep{\fill}}lrrrrrrrr}
\toprule
& \multicolumn{4}{c}{\textbf{All episodes}} & \multicolumn{4}{c}{\textbf{Mean by task family}} \\
\cmidrule(lr){2-5}\cmidrule(lr){6-9}
\textbf{Model} & \textbf{Mean} & \textbf{Median} & \textbf{P90} & \textbf{Std.} & \textbf{Passive} & \textbf{Dynamic} & \textbf{Interaction} & \textbf{Exp. Gen.} \\
\midrule
Qwen3-VL-4B-Ins & 5.57 & 8 & 8 & 2.71 & 5.37 & 5.20 & 6.96 & 6.68 \\
Qwen3-VL-8B-Ins & 5.63 & 8 & 8 & 2.73 & 5.21 & 5.40 & 7.52 & 6.54 \\
Qwen3-VL-32B-Ins & 5.21 & 5 & 8 & 2.53 & 4.35 & 6.02 & 5.81 & 4.65 \\
InternVL3-8B & 6.77 & 8 & 8 & 2.33 & 6.88 & 6.22 & 7.78 & 7.67 \\
InternVL3-38B & 5.32 & 5 & 8 & 2.53 & 4.80 & 6.06 & 5.79 & 3.60 \\
Ovis2-16B & 6.04 & 8 & 8 & 2.57 & 5.37 & 6.32 & 6.92 & 6.92 \\
Qwen3.6-27B & 4.58 & 4 & 8 & 2.54 & 4.39 & 4.28 & 6.48 & 4.65 \\
Mistral-Small-3.1-24B & 5.28 & 5 & 8 & 2.62 & 4.21 & 5.88 & 7.62 & 4.68 \\
GPT-5.4-mini & 5.12 & 5 & 8 & 2.64 & 4.23 & 5.29 & 7.48 & 5.73 \\
GPT-5.4 & 3.96 & 3 & 8 & 2.26 & 3.62 & 3.69 & 6.19 & 4.24 \\
Gemini-2.5-Pro & 3.18 & 2 & 7 & 1.89 & 3.17 & 2.35 & 5.94 & 3.97 \\
Gemini-3-Flash & 3.07 & 2 & 6 & 1.77 & 3.07 & 2.28 & 5.46 & 4.11 \\
\bottomrule
\end{tabular*}
\caption{Recorded output-trajectory length for full-context MLLMs. Overall statistics use all 2,554 episodes per model; family columns report means. P90 is the empirical 90th percentile, Std. is the population standard deviation, and the maximum is eight for every model.}
\label{tab:model-output-trajectory-length}
\end{table}

Table~\ref{tab:model-output-trajectory-length} shows that mean trajectory length ranges from 3.07 actions for Gemini-3-Flash to 6.77 for InternVL3-8B. Ten of the 12 models reach the eight-step budget at the 90th percentile. Averaging each family equally across models gives 4.56 actions for Passive Observation, 4.92 for Dynamic Tracking, 6.66 for Interaction Failure, and 5.29 for Experience Generalization. The corresponding strict-oracle trajectories average 3.00, 3.00, 6.00, and 4.22 actions, respectively, so the longer Interaction Failure trajectories partly reflect its six-action execution structure rather than only model inefficiency. These lengths are descriptive rather than a performance ranking: successful completion can shorten a trajectory, but an incorrect early termination can do so as well, while repeated failed attempts can saturate the budget.

\begin{table}[!t]
\centering
\small
\setlength{\tabcolsep}{4.2pt}
\begin{tabular*}{\textwidth}{@{\extracolsep{\fill}}lrrrrrrr}
\toprule
\textbf{Task family} & \textbf{Episodes} & \textbf{AI2-THOR} & \textbf{ProcTHOR} & \textbf{Scenes} & \textbf{Rooms} & \textbf{Targets} & \textbf{Recept.} \\
\midrule
Passive Observation & 1,036 & 111 & 925 & 1,109 & 4 & 52 & 20 \\
Dynamic Tracking & 1,052 & 109 & 943 & 1,116 & 4 & 54 & 20 \\
Interaction Failure & 263 & 96 & 167 & 423 & 4 & 18 & 5 \\
Experience Generalization & 203 & 36 & 167 & 746 & 4 & 47 & 22 \\
\midrule
Global union & 2,554 & 352 & 2,202 & 1,118 & 4 & 83 & 33 \\
\bottomrule
\end{tabular*}
\caption{Composition of EMem-Bench. AI2-THOR and ProcTHOR assign each episode by the simulator of its target-task scene. Scenes, room types, target object types, and receptacle types are unique within each row; the global-union row deduplicates overlaps across task families.}
\label{tab:benchmark-composition}
\end{table}

\begin{table}[!t]
\centering
\small
\setlength{\tabcolsep}{4.0pt}
\begin{tabular*}{\textwidth}{@{\extracolsep{\fill}}lccccc}
\toprule
\textbf{Task family} & \textbf{Sessions} & \textbf{History steps} & \textbf{Cue gap} & \textbf{Transitions} & \textbf{Visible at cue} \\
& \multicolumn{4}{c}{\textit{Mean / median / 90th percentile}} & \textit{Mean} \\
\midrule
Passive Observation & 2.0 / 2 / 2 & 76.5 / 81.5 / 98 & 49.8 / 56 / 68 & 2.0 / 2 / 2 & 11.0 \\
Dynamic Tracking & 2.0 / 2 / 2 & 78.2 / 83 / 100 & 46.4 / 52 / 75 & 2.0 / 2 / 2 & 9.1 \\
Interaction Failure & 2.0 / 2 / 2 & 71.7 / 78 / 90 & 47.7 / 55 / 64 & 2.0 / 2 / 2 & 6.2 \\
Experience Generalization & 4.5 / 4 / 5 & 112.5 / 111 / 136 & 52.2 / 53 / 62 & 4.5 / 4 / 5 & 9.9 \\
\bottomrule
\end{tabular*}
\caption{Interaction-history difficulty by task family. Cue gap counts context steps after the final decisive evidence and before the target task. Transitions count adjacent scene changes across the context sessions and target scene. Visible at cue is the mean number of visible objects at the decisive evidence step.}
\label{tab:benchmark-difficulty-statistics}
\end{table}

\Needspace{4\baselineskip}
\subsection{Composition and Interaction-History Difficulty}

\begin{wrapfigure}{R}{0.45\textwidth}
\vspace{0pt}
\centering
\includegraphics[width=\linewidth]{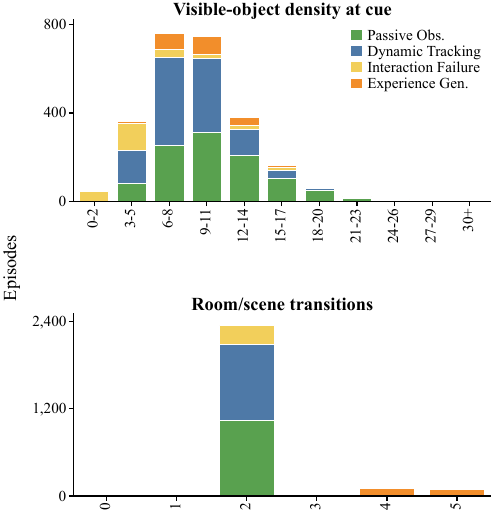}
\caption{Auxiliary benchmark distributions. Top: visible-object density at the decisive cue. Bottom: room/scene transitions across the interaction history. Bars stack episode counts from the four task families.}
\label{fig:benchmark-auxiliary-distributions}
\end{wrapfigure}

Table~\ref{tab:benchmark-composition} details the composition of each task family. Passive Observation and Dynamic Tracking are the two largest families and together account for 81.8\% of the episodes. Interaction Failure and Experience Generalization are smaller controlled sets, but still span 423 and 746 scenes, respectively. By target-scene source, the benchmark contains 352 AI2-THOR episodes and 2,202 ProcTHOR episodes.

The benchmark's memory demand combines target-task execution with a long interaction history. Table~\ref{tab:benchmark-difficulty-statistics} quantifies interaction-history difficulty, and Figure~\ref{fig:benchmark-auxiliary-distributions} reports visible-object density at the decisive cue and room/scene transitions. Experience Generalization is structurally the longest family, with a median of 111 interaction steps and four to five room/scene transitions. The other families have median histories of 78--83 steps and exactly two transitions, yet their median cue-to-target gaps remain comparably long at 52--56 steps. The cue itself is also visually nontrivial: Passive Observation contains 11.0 visible objects on average at the decisive cue, consistent with its fine-grained visual-memory focus. Interaction Failure has fewer visible objects because its decisive state is revealed by action feedback rather than appearance. Together with the main-paper results and the cue-removal controls in Section~\ref{app:additional-experiments}, these statistics show that successful tasks require recovering decisive evidence across long, changing, and visually cluttered interaction histories.

\subsection{Object Coverage and Distribution Concentration}

Table~\ref{tab:benchmark-coverage} summarizes category breadth and concentration, complementing the global unions in Table~\ref{tab:benchmark-composition}. Target-object concentration is low in Passive Observation, Dynamic Tracking, and Experience Generalization: their most frequent target accounts for only 11.0\%, 9.6\%, and 5.9\% of family episodes. Interaction Failure is intentionally narrower, with 18 target types and five receptacle types, so its top-five shares are correspondingly higher. Receptacle distributions are more concentrated than target-object distributions because each task family admits only receptacles that support its executable terminal action. Overall, EMem-Bench covers a broad and nontrivially distributed set of objects and scenes while retaining the controlled object--receptacle support required by each task family.

\begin{table}[!t]
\centering
\small
\setlength{\tabcolsep}{4.0pt}
\begin{tabular*}{\textwidth}{@{\extracolsep{\fill}}lrlrrlr}
\toprule
& \multicolumn{3}{c}{\textbf{Target object}} & \multicolumn{3}{c}{\textbf{Target receptacle}} \\
\cmidrule(lr){2-4}\cmidrule(lr){5-7}
\textbf{Task family} & \textbf{Types} & \textbf{Top-1 (share)} & \textbf{Top-5} & \textbf{Types} & \textbf{Top-1 (share)} & \textbf{Top-5} \\
\midrule
Passive Observation & 52 & Book (11.0\%) & 32.3\% & 20 & Dresser (27.2\%) & 77.0\% \\
Dynamic Tracking & 54 & Book (9.6\%) & 32.0\% & 20 & CounterTop (24.0\%) & 66.0\% \\
Interaction Failure & 18 & KeyChain (30.8\%) & 82.5\% & 5 & Safe (33.8\%) & 100.0\% \\
Experience Generalization & 47 & Candle (5.9\%) & 25.6\% & 22 & Dresser (15.8\%) & 55.2\% \\
\bottomrule
\end{tabular*}
\caption{Category coverage and concentration within each task family. Top-1 and Top-5 are shares of family episodes accounted for by the most frequent one or five values in the indicated scope.}
\label{tab:benchmark-coverage}
\end{table}

\subsection{Release Audit and Executability}

We audit every retained episode before release. All 2,554 episodes belong to the canonical manifest, contain recoverable decisive evidence, execute successfully under the benchmark's strict-oracle action contract, and reach the intended terminal task state. These automatic checks are paired with the six-dimensional manual review below, which covers evidence sufficiency, instruction clarity, answer or target leakage, trajectory plausibility, cue--probe alignment, and terminal-judgment correctness. Together, the construction and review pipeline rejected 143 candidates before retaining the final 2,554 episodes.

\subsubsection{Human Review Protocol and Audit Interface}

\begin{figure}[!htbp]
\centering
\includegraphics[width=0.99\textwidth]{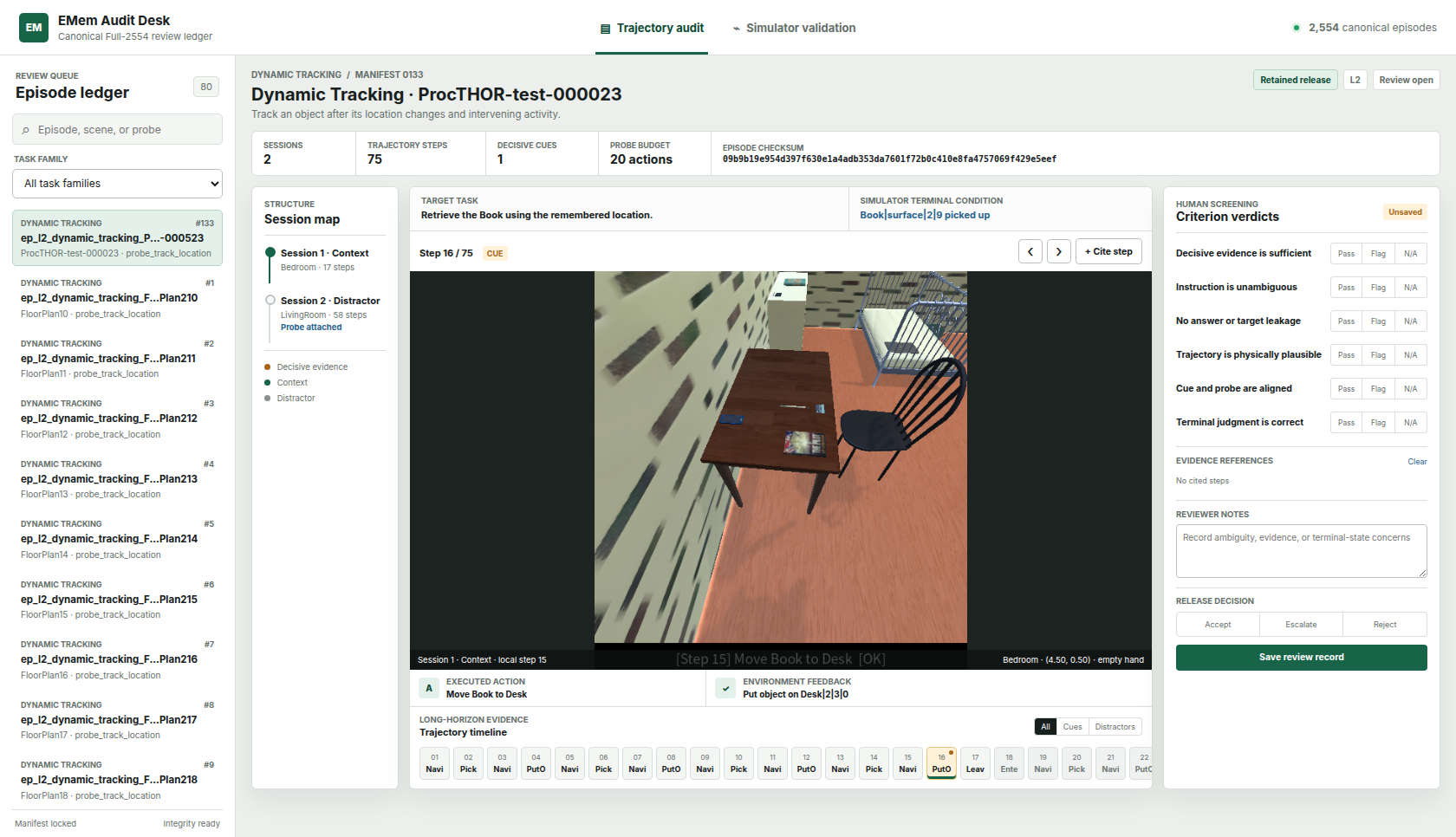}
\caption{EMem-Bench human-audit interface on a canonical Dynamic Tracking episode. The left queue locates an episode in the frozen manifest and the session map preserves context/noise boundaries. The center aligns the target and terminal condition with an enlarged observation, action--feedback pair, and the complete 75-step timeline; the highlighted step records the Book's move from its old Box location to the Desk. The right panel records criterion-level verdicts, cited evidence steps, notes, and the release decision. The \textsc{Simulator validation} mode uses the same workspace to issue benchmark-legal actions and inspect live task progress and terminal-state judgments.}
\label{fig:human-audit-interface}
\end{figure}

\paragraph{Review unit and evidence traversal.}
A panel of five domain experts performed manual screening at the candidate-episode level after the schema, duplication, cue-recovery, and strict-oracle gates and before admission to the frozen manifest. The reviewer first compared the probe instruction and simulator-defined terminal condition, and then traversed the complete context trajectory in chronological order. Figure~\ref{fig:human-audit-interface} shows the benchmark-specific interface used to make long trajectories tractable. Its session map preserves scene and session boundaries; the global timeline retains every step and distinguishes context, distractor, and decisive-evidence steps; and the enlarged view aligns the RGB observation with the executed action, simulator feedback, visible-object list, agent pose, and held object. A construction-metadata highlight anchors the review to the task-defining cue. Reviewers cited global step indices to ground each concern or acceptance decision in recoverable evidence.

\paragraph{Manual acceptance criteria.}
The manual review considered the six dimensions shown in the right panel of Figure~\ref{fig:human-audit-interface}. \emph{Evidence sufficiency} checked that the decisive object, state change, interaction outcome, or repeated household correction was recoverable from the recorded observation--action--feedback chain. \emph{Instruction clarity} checked that the target task and its intended completion state were identifiable. \emph{No answer or target leakage} checked that the later instruction and distractor content preserved dependence on the remembered location, outcome, or generalized rule established by the trajectory. \emph{Trajectory plausibility} checked the consistency of scene transitions, executed actions, environment feedback, held-object state, and relevant object-state changes. \emph{Cue--probe alignment} checked that the probe depended on the object instance, updated state, failed interaction, or experience pattern established by the earlier cue. Finally, \emph{terminal-judgment correctness} checked the evaluator's transition from the pending state to the intended completion state. Candidates with a material issue were corrected and rechecked or removed. This construction-and-screening process rejected 143 candidates and retained the 2,554 episodes reported in the main paper.

\paragraph{Interactive executability and completion audit.}
The second interface mode initializes the candidate's probe through the same \texttt{EmbodiedMemorizerEnv}, legal-action generator, action executor, and \texttt{EpisodeEvaluator} used for benchmark evaluation. It presents the actions legal in the current simulator state, refreshes the RGB observation and object/agent state after every transition, and displays the evaluator's task progress and \texttt{task\_completed} value. The reviewer can replay the declared oracle path, inspect every action result, and reset a suspicious case to test an alternative legal path. This interactive audit establishes task executability and verifies the evaluator's transition from pending progress to the intended terminal state.

\Needspace{4\baselineskip}
\subsection{Task-Family Sensitivity}

\begin{wrapfigure}{R}{0.45\textwidth}
\vspace{0pt}
\centering
\includegraphics[width=\linewidth]{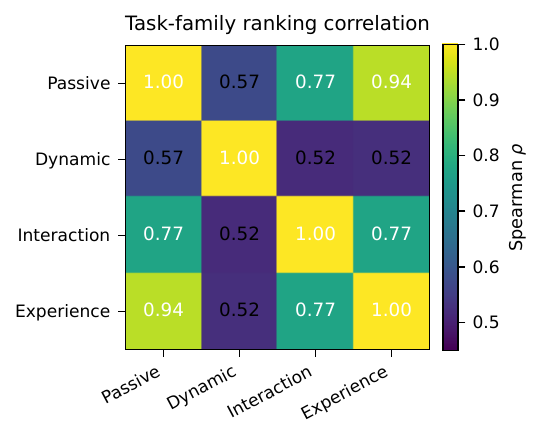}
\caption{Spearman correlations between model SR rankings across task families, computed from the 12 full-context MLLMs that completed all 2,554 episodes.}
\label{fig:benchmark-evaluation-sensitivity}
\end{wrapfigure}

Figure~\ref{fig:benchmark-evaluation-sensitivity} compares the task-family SR rankings of the 12 open-source and proprietary full-context MLLMs that completed all 2,554 episodes. Pairwise correlations range from 0.52 to 0.94. Passive Observation and Experience Generalization are most aligned, whereas Dynamic Tracking has only 0.52 correlation with Interaction Failure and Experience Generalization. The positive but nonuniform correlations match the main result: overall model strength transfers across families, but models retain uneven profiles because the families exercise different uses of memory.

\subsection{Case Studies}
\label{app:benchmark-case-studies}

\begin{figure}[p]
\centering
\includegraphics[width=0.99\textwidth,height=0.86\textheight,keepaspectratio]{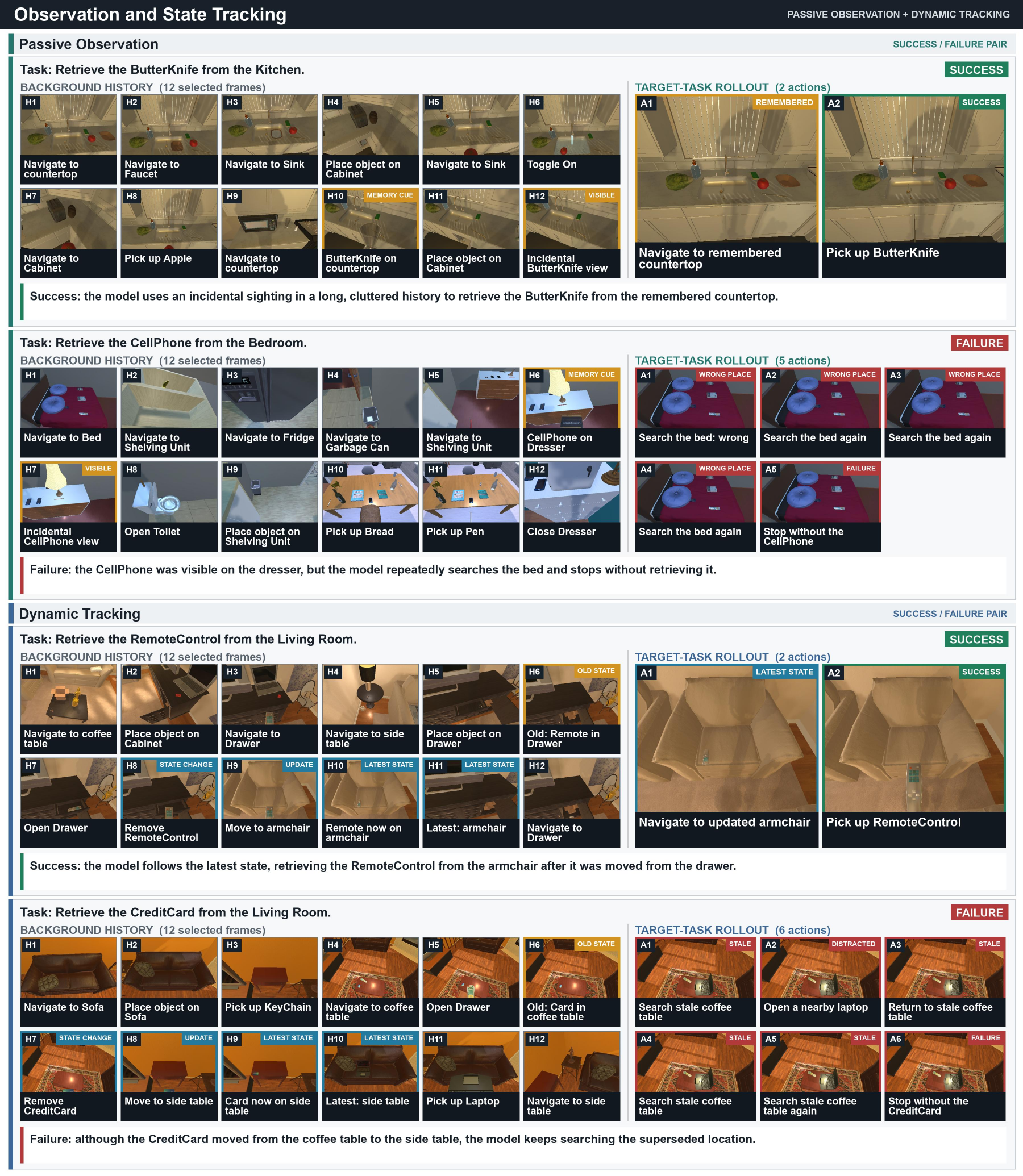}
\caption{Passive Observation and Dynamic Tracking cases. The Passive Observation pair contrasts retrieval from an incidental sighting with repeated search at an incorrect location. The Dynamic Tracking pair contrasts use of the latest object location with repeated reliance on a superseded state. Colored borders identify the task-defining historical cues and consequential rollout actions.}
\label{fig:case-observation-tracking}
\end{figure}

\begin{figure}[p]
\centering
\includegraphics[width=0.99\textwidth,height=0.86\textheight,keepaspectratio]{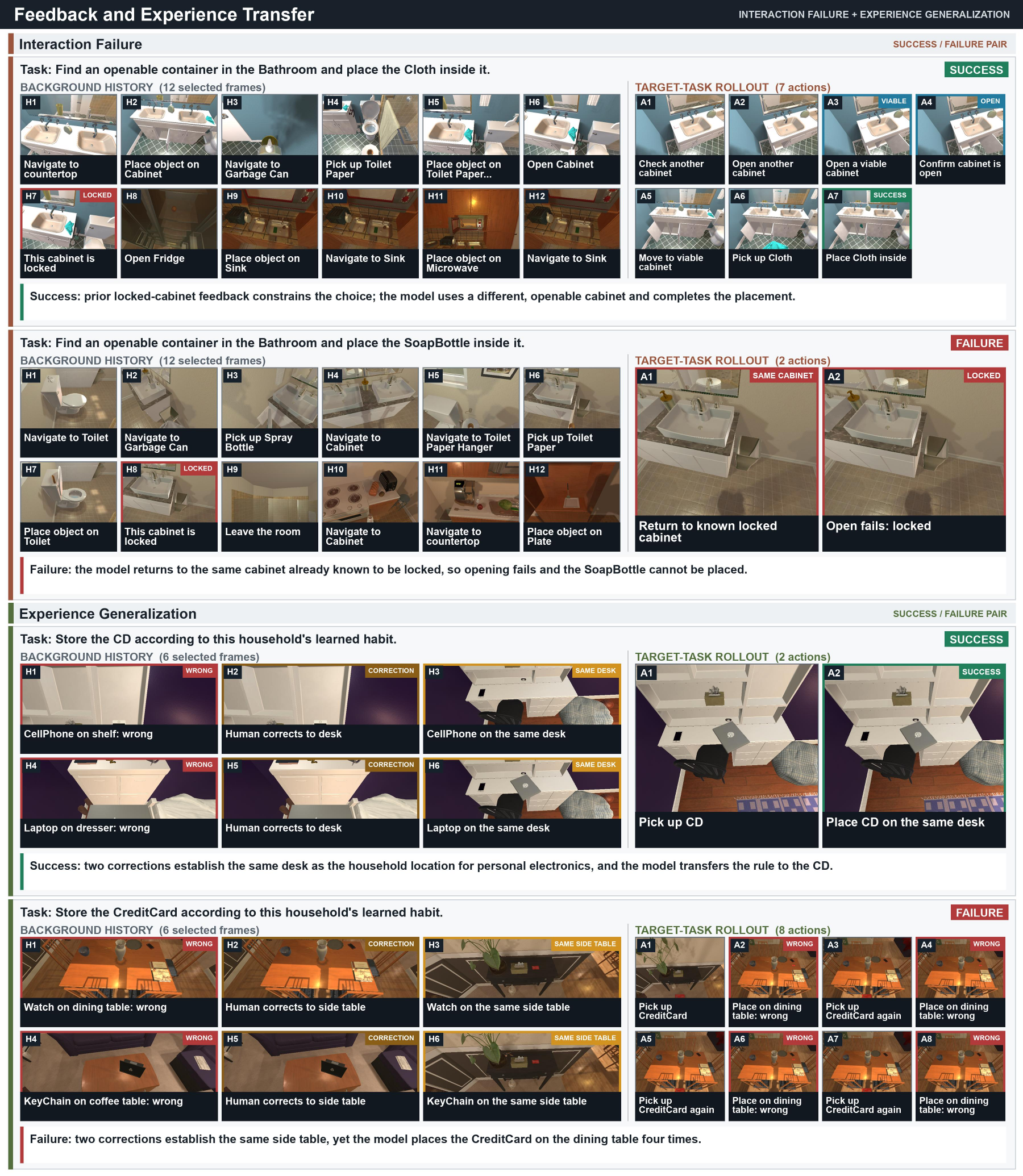}
\caption{Interaction Failure and Experience Generalization cases. The Interaction Failure pair contrasts avoiding a cabinet known to be locked with returning to that same unusable instance. The Experience Generalization pair contrasts transferring two same-surface corrections to a novel object with repeatedly violating the learned convention. Each generalization case remains within one physical scene and one shared target surface.}
\label{fig:case-feedback-generalization}
\end{figure}

Figures~\ref{fig:case-observation-tracking} and~\ref{fig:case-feedback-generalization} provide paired success and failure examples from GPT-5.4-mini under the \texttt{full\_context} protocol. Each background panel samples the underlying interaction trajectory in chronological order while retaining every task-defining cue; each rollout panel shows the complete target-task action trace used in the comparison. The background RGB frames visualize the recorded history for the reader.

\mypar{Passive Observation.}
Figure~\ref{fig:case-observation-tracking} (top) isolates whether fine-grained visual evidence survives a long, distracting history. In the success case, the ButterKnife is incidentally visible among many countertop objects, and the model later navigates to that remembered location and retrieves it. In the failure case, the history similarly makes the CellPhone recoverable on the dresser, but the model repeatedly searches the bed and terminates without the target. The comparison isolates the later policy's use of recoverable historical evidence.

\mypar{Dynamic Tracking.}
Figure~\ref{fig:case-observation-tracking} (bottom) contrasts use of the latest state with retention of a superseded state. In the success case, the history moves the RemoteControl from a drawer to an armchair, and the model retrieves it from the armchair. In the failure case, the CreditCard moves from the coffee table to the side table, yet the model repeatedly returns to the coffee table and stops. This pair realizes the main-paper distinction between merely retaining an earlier observation and replacing it with a later world-state update.

\mypar{Interaction Failure.}
Figure~\ref{fig:case-feedback-generalization} (top) shows how a constraint revealed by interaction feedback affects a later choice. Both histories report a locked cabinet. The success case avoids the known unusable instance, opens a different cabinet, and places the Cloth inside it. The failure case instead returns to the same known locked cabinet; its opening attempt fails while the cabinet remains closed, preventing placement of the SoapBottle. Successful transfer associates the earlier feedback with the identity of the later interaction target.

\mypar{Experience Generalization.}
Figure~\ref{fig:case-feedback-generalization} (bottom) grounds each pair of corrections and its probe in one physical scene and one shared target surface. In the success case, corrections place a CellPhone and a Laptop on the same desk; the model transfers this recurring household convention to a new CD. In the failure case, corrections instead establish the same side table for a Watch and a KeyChain, but the model places the novel CreditCard on the dining table four times. The contrast tests induction of a reusable rule from multiple corrected experiences.

\section{Additional Experimental Results}
\label{app:additional-experiments}

We conduct two controlled experiments that complement the broad comparison in Table~\ref{tab:leaderboard}. They test whether success changes causally with episode evidence and whether adding historical images to the main-paper \texttt{full\_context} setting affects task success. Both experiments use the same fixed episode identifiers released with the evaluation ledger. Average SR is the arithmetic mean of the four family SRs, as in the main paper. We report episode-paired bootstrap 95\% confidence intervals from 10,000 resamples; paired differences always reuse the same resampled episode identifiers.

\subsection{Causal Dependence on Interaction History}

We draw a fixed 800-episode subset from the canonical manifest, with 200 episodes per task family and balanced sampling over cue-to-task distance quartiles. Qwen3-VL-8B and GPT-5.4-mini are evaluated on the same episodes under four history conditions. \emph{Correct history} uses the standard \texttt{full\_context} protocol. \emph{Current only} retains the instruction and current observation but removes all preceding sessions. \emph{Cue removed} preserves the session boundaries and distractors while replacing the decisive evidence record with a length- and type-matched neutral record that does not determine the target relation. \emph{Minimal oracle} removes the remaining history and supplies only the shortest structured proposition sufficient to identify the intended terminal relation.

Table~\ref{tab:causal-history-controls} reports overall and family-level SR under these four interventions. We define the paired cue effect as
$\Delta_{\mathrm{cue}}=\mathrm{SR}_{\mathrm{correct}}-\mathrm{SR}_{\mathrm{cue\mbox{-}removed}}$.

\begin{table}[!t]
\centering
\small
\setlength{\tabcolsep}{4.0pt}
\begin{tabular*}{\textwidth}{@{\extracolsep{\fill}}llrrrrr}
\toprule
\textbf{Backbone} & \textbf{History condition} & \textbf{Avg} & \shortstack{\textbf{Passive}\\\textbf{Observation}} & \shortstack{\textbf{Dynamic}\\\textbf{Tracking}} & \shortstack{\textbf{Interaction}\\\textbf{Failure}} & \shortstack{\textbf{Experience}\\\textbf{Generalization}} \\
\midrule
\multirow{4}{*}{Qwen3-VL-8B}
& Correct history & 23.8 & 37.5 & 23.5 & 12.5 & 21.5 \\ 
& Current only & 10.8 & 8.5 & 8.5 & 18.0 & 8.0 \\ 
& Cue removed & 12.6 & 8.5 & 23.0 & 12.5 & 6.5 \\ 
& Minimal oracle & 58.2 & 64.0 & 61.5 & 25.0 & 82.5 \\ 
\midrule
\multirow{4}{*}{GPT-5.4-mini}
& Correct history & 36.6 & 58.0 & 28.5 & 19.0 & 41.0 \\ 
& Current only & 11.0 & 8.0 & 11.5 & 19.0 & 5.5 \\ 
& Cue removed & 15.2 & 11.0 & 15.0 & 20.5 & 14.5 \\ 
& Minimal oracle & 69.1 & 85.0 & 85.5 & 48.0 & 58.0 \\ 
\bottomrule
\end{tabular*}
\caption{History interventions for Qwen3-VL-8B and GPT-5.4-mini. All four conditions use the same 800 episodes, with 200 episodes from each task family. Entries are SR percentages. We compute $\Delta_{\mathrm{cue}}$ directly from the paired Correct history and Cue removed outcomes and report its bootstrap 95\% confidence interval in the text.}
\label{tab:causal-history-controls}
\end{table}

The paired Correct-history minus Cue-removed macro-SR difference was 11.1 points for Qwen3-VL-8B (95\% CI [8.1, 14.1]) and 21.4 points for GPT-5.4-mini (95\% CI [17.8, 25.0]). Across cue-distance quartiles, the corresponding paired differences were Q1: 14.5, Q2: 10.5, Q3: 5.5, Q4: 14.0 points for Qwen3-VL-8B and Q1: 25.0, Q2: 18.0, Q3: 21.0, Q4: 21.5 points for GPT-5.4-mini. Taken together, these interventions establish the central validity claim of EMem-Bench: benchmark success genuinely requires memory of preceding interactions. With the target task and current environment state held fixed, removing the earlier sessions or only the decisive historical evidence sharply reduces macro SR; the tasks therefore cannot be solved from the current observation, generic scene priors, or ordinary embodied competence alone. The positive cue effect in every distance quartile confirms that this requirement extends to temporally remote evidence. Minimal oracle reaches 58.2 and 69.1 macro SR for Qwen3-VL-8B and GPT-5.4-mini, exceeding Current only by 47.4 and 58.1 points, respectively. It also exceeds Correct history by 34.4 and 32.5 points. These gaps show that both models can act much more effectively once the task-relevant past relation is recovered, separating the memory-retrieval challenge from downstream execution. Memory is therefore necessary information for solving EMem-Bench tasks.

\subsection{Historical Visual Inputs in Full Context}
\label{app:historical-visual-input}

This section expands the context-extension experiment in Section~\ref{sec:visual-history-expansion}; Table~\ref{tab:full-context-visual-input} reports the detailed task-family results. We compare the main-paper \texttt{full\_context} setting against a maximum-history-image variant. GPT-5.4-mini is evaluated on the same fixed 800 episodes used in the causal controls. The main-paper setting uses the complete sanitized text history and current $500\times500$ RGB observation defined in Section~\ref{app:benchmark-details}. The maximum-history-image setting preserves this input unchanged and fills the remaining model context with as many historical $500\times500$ RGB observations as the provider's input limit permits after reserving a fixed output budget.

\begin{table}[!tbp]
\centering
\small
\setlength{\tabcolsep}{4.0pt}
\begin{tabular*}{\textwidth}{@{\extracolsep{\fill}}lrrrrrrr}
\toprule
\textbf{Input setting} & \textbf{Avg} & \shortstack{\textbf{Passive}\\\textbf{Observation}} & \shortstack{\textbf{Dynamic}\\\textbf{Tracking}} & \shortstack{\textbf{Interaction}\\\textbf{Failure}} & \shortstack{\textbf{Experience}\\\textbf{Generalization}} & \textbf{Input tok.} & \textbf{s/step} \\
\midrule
Main-paper \texttt{full\_context} & 36.6 & 58.0 & 28.5 & 19.0 & 41.0 & 41{,}770 & 4.1 \\ 
\shortstack[l]{Main-paper \texttt{full\_context}\\+ maximum historical RGB} & 26.2 & 40.0 & 21.0 & 20.0 & 24.0 & 206{,}489 & 39.1 \\ 
\bottomrule
\end{tabular*}
\caption{Main-paper versus maximum-history-image \texttt{full\_context} settings for GPT-5.4-mini on the same 800-episode main-selector subset used by the four history controls in Table~\ref{tab:causal-history-controls}. Both conditions contain the complete sanitized text history and the current RGB observation at every target-task step; the second condition additionally includes as many historical RGB observations as the remaining model context permits. Average and task-family SRs are percentages. Input tokens are per-episode means, and wall-clock latency per model step excludes simulator rendering. We report the episode-paired macro-SR change and its bootstrap 95\% confidence interval in the text.}
\label{tab:full-context-visual-input}
\end{table}

When the complete visual history exceeds the provider limit, a deterministic sampler allocates frames uniformly across sessions and each session's timeline, then restores them to temporal order. The resulting treatment augments the main-paper input with the maximum historical RGB sequence admitted by the model context. Image-token counts are computed with the model's image processor.

Adding the maximum available historical RGB sequence changed macro SR by -10.4 points (95\% CI [-14.4, -6.5]). The treatment admitted archived frames for all 800 episodes. The paired SR changes across history-length quartiles were Q1: -11.5, Q2: -8.5, Q3: -9.0, Q4: -12.5 points. Taken together, these results establish a broader conclusion: EMem-Bench cannot be solved by simply enlarging the raw multimodal context. Maximizing historical visual input reduced rather than improved macro SR, while increasing mean input from 41{,}770 to 206{,}489 tokens per episode and mean model-step latency from 4.1 to 39.1 seconds. The negative change in every history-length quartile shows that the degradation is systematic across history lengths. Additional observations are therefore insufficient for effective memory. Success requires precise retrieval of task-relevant evidence and suppression of irrelevant history; indiscriminate context accumulation instead dilutes the decisive signal. EMem-Bench consequently distinguishes long-term memory retrieval capability from mere context-window scaling.

\section{Limitations}
A limitation of this work is that EMem-Bench and the cross-benchmark evaluations are conducted primarily in simulation. Simulation allows us to control the earlier evidence, reproduce complete interactions, and score the resulting environment transitions while reducing the cost and safety risks of physical experiments. However, it does not capture sensor noise, actuation error, open-world changes, or ambiguous human feedback encountered in real deployments. The reported results therefore do not establish long-term memory performance on physical robots. Future work will extend the evaluation to longer time horizons and more realistic environments, including standardized real-robot test suites that preserve reproducibility while remaining practical and safe.

\end{document}